\documentclass[journal,twoside]{IEEEtran}
\usepackage{graphicx}
\usepackage{color,soul}
\usepackage{multirow,booktabs,bigstrut,rotating,threeparttable}
\usepackage{cite}

\newcommand{\etal}{\textit{et al.}}
\usepackage{amsmath}
\usepackage{amsfonts}
\usepackage{siunitx}
\usepackage{cases}
\usepackage{bm}
\usepackage{algorithm}
\usepackage{algorithmic}
\usepackage{balance}

\usepackage{array}
\usepackage{makecell}

\ifCLASSOPTIONcompsoc
  \usepackage[caption=false,font=normalsize,labelfont=sf,textfont=sf]{subfig}
\else
  \usepackage[caption=false,font=footnotesize]{subfig}
\fi
 \let\MYorigsubfloat\subfloat
 \renewcommand{\subfloat}[2][\relax]{\MYorigsubfloat[]{#2}}
\usepackage{url}

\newcommand{\mathopr}[1]{\mathtt{#1}}

\usepackage{tikz,xcolor,hyperref}
\hypersetup{hidelinks}

\definecolor{lime}{HTML}{A6CE39}
\DeclareRobustCommand{\orcidicon}{%
	\begin{tikzpicture}
	\draw[lime, fill=lime] (0,0) 
	circle [radius=0.16] 
	node[white] {{\fontfamily{qag}\selectfont \tiny ID}};
	\draw[white, fill=white] (-0.0625,0.095) 
	circle [radius=0.007];
	\end{tikzpicture}
	\hspace{-3mm}
}

\foreach \x in {A, ..., Z}{%
	\expandafter\xdef\csname orcid\x\endcsname{\noexpand\href{https://orcid.org/\csname orcidauthor\x\endcsname}{\noexpand\orcidicon}}
}

\begin{document}
%
\title{Lightweight Stepless Super-Resolution of Remote Sensing Images via Saliency-Aware Dynamic Routing Strategy}
%
%

\author{Hanlin~Wu\orcidA{},~\IEEEmembership{Graduate Student Member,~IEEE},~Ning~Ni\orcidB{},~Libao Zhang\orcidB{},~\IEEEmembership{Member,~IEEE}
    \thanks{
        This work was supported in part by the Beijing Natural Science Foundation under Grant 4222046, in part by the National Natural Science Foundation of China under Grant 62271060, Grant 61571050 and Grant 41771407.
        \textit{(Corresponding author: Libao Zhang.)}

        The authors are with the School of Artificial Intelligence, Beijing Normal University, Beijing 100875, China. (e-mail: libaozhang@bnu.edu.cn).}
}

%
%

\markboth{}%
{}
%



\maketitle

\begin{abstract}
    Deep learning-based algorithms have greatly improved the performance of remote sensing image (RSI) super-resolution (SR). However, increasing network depth and parameters cause a huge burden of computing and storage. Directly reducing the depth or width of existing models results in a large performance drop. We observe that the SR difficulty of different regions in an RSI varies greatly, and existing methods use the same deep network to process all regions in an image, resulting in a waste of computing resources. In addition, existing SR methods generally predefine integer scale factors and cannot perform stepless SR, i.e., a single model can deal with any potential scale factor. Retraining the model on each scale factor wastes considerable computing resources and model storage space. To address the above problems, we propose a saliency-aware dynamic routing network (SalDRN) for lightweight and stepless SR of RSIs. First, we introduce visual saliency as an indicator of region-level SR difficulty and integrate a lightweight saliency detector into the SalDRN to capture pixel-level visual characteristics. Then, we devise a saliency-aware dynamic routing strategy that employs path selection switches to adaptively select feature extraction paths of appropriate depth according to the SR difficulty of sub-image patches. Finally, we propose a novel lightweight stepless upsampling module whose core is an implicit feature function for realizing mapping from low-resolution feature space to high-resolution feature space. Comprehensive experiments verify that the SalDRN can achieve a good trade-off between performance and complexity. The code is available at \url{https://github.com/hanlinwu/SalDRN}.
\end{abstract}



\ifCLASSOPTIONpeerreview
    \begin{center} \bfseries EDICS Category: 3-BBND \end{center}
\fi
%
\IEEEpeerreviewmaketitle

\section{Introduction}
%
%
%
%

\IEEEPARstart{S}{ingle} image super-resolution (SISR) aims to recover a high-resolution (HR) image with clear details from its low-resolution (LR) version. SISR is a basic task of remote sensing image (RSI) processing and is helpful for subsequent interpretation tasks such as scene classification \cite{wang2018scene}, fine-scale land cover classification \cite{he2022generating}, object recognition \cite{wang2020rsnet}, and change detection \cite{shi2021deeply}. With the rapid development of deep learning technology, SISR methods based on convolutional neural networks (CNNs) have made significant progress in recent years.

\begin{figure}[t]
    \centering
    \includegraphics[width=\linewidth]{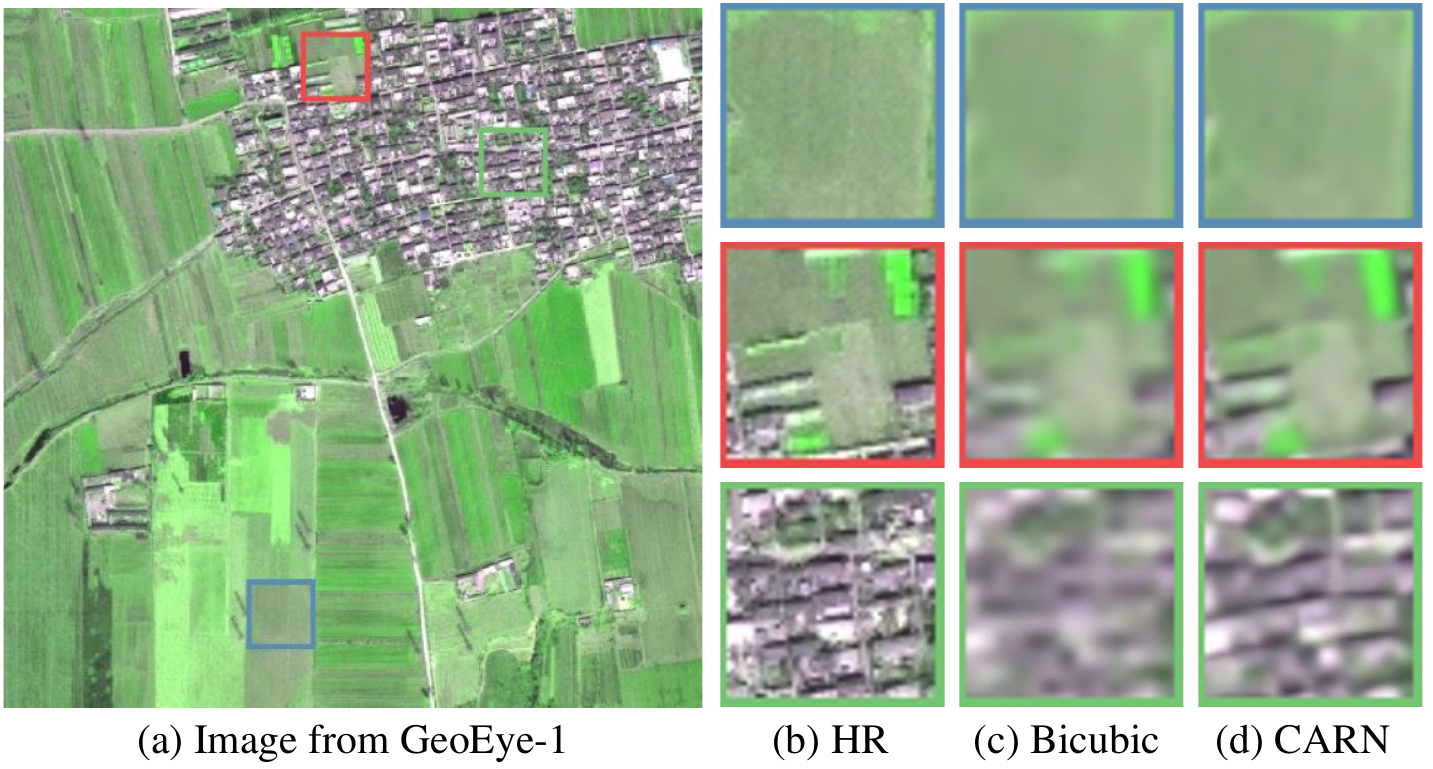}
    \caption{SR difficulty of different regions in RSIs. (a) An RSI from the \mbox{GeoEye-1} satellite; (b) HR image patches; (c) Bicubic interpolation results; (d) SR results of CARN.}
    \label{fig:intro}
\end{figure}

Early CNN-based SISR algorithms built shallow neural networks to learn the mapping from LR to HR images. With the progress of research, although the performance of models has been dramatically improved, the network depth has increased, as has the parameter quantity and computational complexity. For example, the super-resolution convolutional neural network (SRCNN) \cite{dong2015image} proposed in 2015 contains only three layers and $57\,\mathrm{K}$ parameters, the enhanced deep super-resolution network (EDSR) \cite{lim2017enhanced} proposed in 2017 contains $68$ layers and $43\,\mathrm{M}$ parameters, the residual dense network (RDN) \cite{zhang2018residual} proposed in 2018 contains $150$ layers and $23\,\mathrm{M}$ parameters, and the state-of-the-art (SOTA) transformer-based SR network \cite{chen2021pre} has more than $115\,\mathrm{M}$ parameters. The ever-increasing computational complexity consumes an increasing number of computing resources, requiring models to be deployed on high-performance computing platforms.

RSIs have the characteristics of large image size and massive data and face a contradiction between high-speed data acquisition and low-speed interpretation. The traditional RSI processing mode transmits massive data back to the ground and then interprets it, which requires high data transmission bandwidth and cost. Direct on-satellite image processing becomes a pressing need because of the ability to respond quickly to user needs and provide real-time feedback. However, the storage and computing capabilities of on-board devices are usually limited, which puts forward new requirements for real-time and high-efficiency algorithms. Therefore, improving the efficiency of the algorithm while maintaining performance is an essential social requirement in the application of remote sensing. Furthermore, SR is an important preprocessing step that helps improve the performance of subsequent tasks. To apply the RSI SR methods to a broader range of practical scenarios such as on-board computing, it is necessary to build lightweight RSI SR models. 

Most existing SR algorithms were designed to deal with a single and fixed scale factor. Stepless SR aims to super-resolve images with arbitrary \emph{integer/non-integer} scale factors using a \emph{single} model. Building a stepless SR model that can deal with multiple scale factors simultaneously has attracted researchers' interest because it can dramatically save model training and storage resources. Hu \etal \cite{hu2019meta} proposed a meta-learning-based scale-arbitrary SR (Meta-SR) method, which is a pioneering work on the stepless SR task. Subsequently, Chen \etal \cite{chen2021learning} introduced local implicit image functions (LIIFs) to improve the performance of stepless SR. However, both of the above models are designed based on heavyweight networks. Behjati \etal \cite{behjati2021overnet} proposed an overscaling network (OverNet) to reduce the complexity of stepless SR. However, OverNet still has too many parameters, and there is a certain gap in performance compared with the previous fixed-scale SR models. Therefore, it is necessary to design a new stepless upsampling scheme to meet the needs of lightweight SR models.

Previous lightweight SR models \cite{dong2016accelerating,shi2016real,lai2017deep,ahn2018fast,hui2019lightweight} mainly focused on designing a lightweight network structure. They use the same deep network to process all regions in the image, ignoring the fact that the SR difficulty of different regions in an image varies greatly, resulting in a waste of computing resources. We observe that in RSIs, regions with smooth textures are very easy to super-resolve, and even simple bicubic interpolation can yield satisfactory results. For regions with complex textures, deeper networks are required to obtain good SR results. Fig.\,\ref{fig:intro}\,(a) shows an RSI from the GeoEye-1 satellite, and Fig.\,\ref{fig:intro}\,(b)-(d) are HR image patches, bicubic interpolation results, and SR results of the cascading residual network (CARN) \cite{ahn2018fast}, respectively. The first row of Fig.\,\ref{fig:intro} shows a smooth region, and it can be seen that the result of bicubic interpolation is similar to CARN. The last row of Fig.\,\ref{fig:intro} shows a textured region, where bicubic interpolation can hardly recover texture details, and it needs to rely on deep CARN to obtain sharp details. The SR difficulty of the middle row is moderate. This indicates that if different depth networks can be used for feature extraction in different regions of the image, then the computational complexity of the network will be significantly reduced. To achieve this goal, we introduce visual saliency as a guideline for the SR difficulty of image patches and propose a saliency-aware dynamic routing strategy. 

In this article, we propose a novel saliency-aware dynamic routing network (SalDRN) for lightweight and stepless SR. The proposed SalDRN integrates a lightweight saliency detector and a dynamic routing strategy for efficient feature extraction. We employ visual saliency as a guideline of SR difficulty and choose a feature extraction path of appropriate depth for each sub-image patch, thereby avoiding the waste of computational resources. In addition, we devise a lightweight stepless upsampling module (LSUM) to achieve scale-arbitrary SR within a single model. In the LSUM, we first generate progressive resolution feature maps to approximate a continuous representation of an image via the cascaded frequency enhancement block (CFEB), which uses group convolution and channel splitting to reduce computational complexity. Then, we design an efficient implicit feature function (IFF) and combine it with a scale-aware pixel attention (SAPA) mechanism to learn the mapping from the LR to the HR feature space. SAPA can effectively enhance the expression capacity of IFFs and reduce the computational complexity of the upsampling module.

The main contributions of this article are as follows:
\begin{enumerate}
    \item We propose a saliency-aware dynamic routing strategy that can automatically select a feature extraction path with an appropriate depth according to the SR difficulty of the image patch. This reduces the computational complexity while maintaining the performance.
    \item We propose a novel LSUM that combines CFEB and IFF to efficiently map LR features to HR features with arbitrary scale factors. We also devise an SAPA mechanism to enhance the expression capacity of the IFF.
    \item We propose a novel SalDRN for the lightweight and stepless SR of RSIs. The number of FLOPs of our proposal is only $22\%$ of the CFSRCNN with similar performance. Compared with the SOTA stepless upsampling module LIIF, our proposed LSUM has a $98\%$ reduction of FLOPs and an $80\%$ reduction in the number of parameters, achieving comparable performance. 
\end{enumerate}



\section{Related Works}
\label{sec:related_works}

\subsection{Lightweight SR}

SR is a typical computing-intensive task, so building a lightweight SR model to save computational resources has attracted widespread attention from researchers. 

Dong \etal \cite{dong2016accelerating} and Shi \etal \cite{shi2016real} proposed placing the upsampling module at the end of the network so that feature extraction is performed in the LR space, and the computational complexity can be effectively reduced. Lai \etal \cite{lai2017deep} introduced a deep Laplacian pyramid network to reconstruct HR images at multiple pyramid levels progressively. Ahn \etal \cite{ahn2018fast} introduced the group convolution and proposed the CARN to reduce the computational burden of convolution layers in the feature extraction part. Hui \etal \cite{hui2018fast,hui2019lightweight} proposed an information distillation paradigm using a channel splitting operation to extract multilevel features and then aggregate them to improve the capacity of lightweight networks. Zhao \etal \cite{zhao2020efficient} proposed a pixel attention mechanism to improve the performance of lightweight SR networks. Tai \etal \cite{tai2017image} and Li \etal \cite{li2019feedback} adopt a recurrent structure to reduce the number of parameters by weight sharing, but they cannot reduce the computational complexity in the testing phase. Tian \etal \cite{tian2020coarse} proposed a coarse-to-fine super-resolution convolutional neural network (CFSRCNN) that combined long and short path features to prevent performance degradation, and proposed a feature fusion schema to improve the model efficiency. Chu \etal \cite{chu2021fast} used neural architecture search technology to automatically find an effective feature extraction structure and neuron connections for the SR task. 

The above algorithms have greatly reduced the model complexity, but there is still a large gap between the performance of these lightweight models and the SOTA SR models due to the significant decrease in the number of parameters and computational complexity.

\subsection{Stepless SR}

Most existing SR algorithms need to train and store a model for each scale factor. Considering the model flexibility and the cost of training and storage, research on stepless SR is gaining popularity. 

Meta-SR \cite{Hu2019} is a pioneering work on stepless SR that proposes a meta-upscale module for arbitrary scale factor upsampling. The meta-upscale module projects the coordinates of the HR space into the LR space through a many-to-one mapping and then utilizes a fully connected network to dynamically predict the filter weights to upsample each scale factor. The coordinate projection operation makes a single feature vector in the LR space responsible for generating a small patch of the SR image. Two adjacent pixels in the SR result may not belong to the same image patch, so the discontinuous image representation will cause unpleasant checkerboard artifacts when the LR feature vector is switched. To solve the problem of checkerboard artifacts, Chen \etal \cite{chen2021learning} introduced LIIF and a local ensemble technique to obtain a continuous representation of an image. However, the local ensemble strategy creates an additional computational burden because repeated predictions need to be made with different latent codes. Behjati \etal \cite{behjati2021overnet} proposed an overscaling module (OSM) that uses overscaled feature maps for stepless upsampling. However, OSM also has considerable computational complexity due to the generation of the overscaled feature maps, and it can only perform SR with a scale factor below the predefined maximum value. Wang \etal \cite{wang2021learning} proposed a scale-aware upsampling layer that first achieves asymmetric SR. Xu \etal \cite{xu2021ultrasr} introduced spatial encoding to address the structural distortions problem in the LIIF.

Although the above methods improved the performance of stepless SR, they still face a high computational burden and are challenging to deploy and apply. 

\subsection{Region-Aware SR}

Some researchers adopted different processing strategies for regions with different characteristics in the image. Romano \etal \cite{romano2016raisr} proposed a rapid and accurate image super-resolution (RAISR) algorithm, dividing image patches into clusters and constructing appropriate filters for each cluster. RAISR adopts an efficient hashing algorithm to reduce the complexity of clustering. Wang \etal \cite{wang2018recovering} introduced a spatial feature transformation layer combined with high-level semantic information to provide different priors for different regions. Yu \etal \cite{yu2018crafting,yu2021path} proposed decomposing the input image into sub-patches and implementing network path selection through reinforcement learning. 

Inspired by the human visual attention mechanism \cite{gilbert2007brain}, some researchers introduced saliency analysis to guide computer vision tasks \cite{zhang2014saliency,cehn2020salience} . In the image, the saliency of a region is the state or quality by which it stands out from its neighbors. Ma \etal \cite{ma2020saliency} and Wu \etal \cite{wu2022remote} proposed a saliency-guided feedback generative adversarial network (SG-FBGAN) that uses a saliency map as an indicator of texture complexity, so different reconstruction principles can be applied to restore areas with varying saliency levels. However, SG-FBGAN requires an external saliency detection algorithm and thus cannot achieve end-to-end model training. 

Although these algorithms considered the diverse reconstruction needs of different regions in the image, they did not take advantage of this feature to speed up the model.

\begin{figure*}[t]
    \centering
    \includegraphics[width=\textwidth]{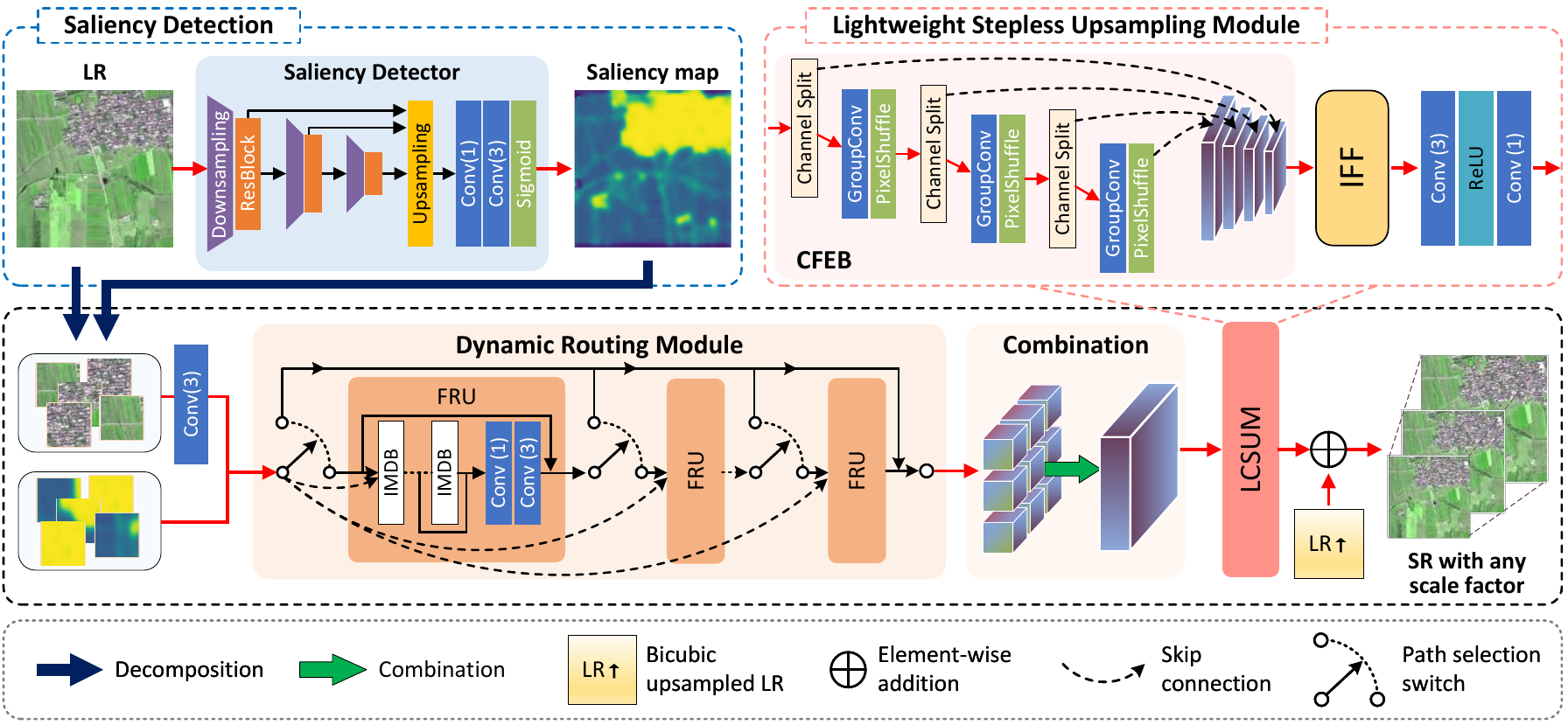}
    \caption{Framework of the proposed SalDRN. The downsampling layer is a $3\times 3$ convolutional layer with a stride of $2$, and upsampling layer is a nearest interpolation followed by a $3\times 3$ convolutional layer; ``Conv(k)'' represent a $k\times k$ convolutional layer. }
    \label{fig:flowchart}
\end{figure*}

\subsection{SR for RSIs}

Most of the early CNN-based SR algorithms for RSIs are improvements of SRCNN \cite{dong2015image} or EDSR \cite{lim2017enhanced}, and various network structures are designed to extract the features of RSIs more effectively. Lei \etal \cite{lei2017super} proposed a local-global combined network designed with a multifork structure to learn multilevel representations of RSIs. Haut \etal \cite{haut2018a} proposed an unsupervised convolutional generative model that learns relationships between the LR and HR domains throughout several convolutional, downsampling, batch normalization, and activation layers. Later, many researchers considered the characteristics of RSIs to design the network structure \cite{pan2019super,lei2019coupled,li2021exploring}. Hou \etal \cite{hou2018adaptive} proposed an effective method for RSI SR based on sparse representation. They introduced a global self-compatibility model for global regularization and improved the performance of the model by integrating the sparse representation and the local and nonlocal constraints. Zhang \etal \cite{zhang2021nonpairwise} proposed a new cycle convolutional neural network (Cycle-CNN), which can be trained with unpaired data. The Cycle-CNN solves the problem in which paired HR and LR RSIs are actually difficult to acquire. Chen \etal \cite{chen2021remote} proposed a residual aggregation and split attentional fusion network that uses a basic split-fusion mechanism to achieve cross-channel feature group interaction, allowing the method to adapt to various land surface scene reconstructions. Shao \etal \cite{shao2019remote} proposed a coupled sparse autoencoder to effectively learn the mapping relation between LR and HR images. Dong \etal \cite{dong2021remote} developed a dense-sampling SR network (DSSR) to explore the large-scale SR of RSIs. DSSR incorporates a wide activation and attention mechanism to enhance the representation ability of the network. Recently, some studies \cite{zhang2020scene,zhang2020remote} introduced the attention mechanism into the RSI SR task to improve the expression ability of networks.

At present, few scholars have considered building a lightweight SR model for RSIs. Wang \etal \cite{wang2021contextual} proposed a lightweight contextual transformation network (CTN) by replacing the classical $3\times 3$ convolutional layer with lightweight contextual transformation layers. Wang \etal \cite{wang2022fenet} proposed a lightweight feature enhancement network for accurate RSI SR. They designed a lightweight lattice block as a nonlinear feature extraction function to improve the expression ability and enabled the upper and lower branches to communicate efficiently by using the attention mechanism.

Research on the stepless SR of RSIs started relatively late. To the best of our knowledge, only the model in \cite{ni2021hierarchical} considers the stepless SR of RSIs. Ni \etal \cite{ni2021hierarchical} constructed a self-learning network by aggregating hierarchical features but did not consider the issue of model lightweighting.
\section{Proposed Method}
\label{sec:methodology}

First, we provide an overview of our proposed SalDRN in Sec.\,\ref{sec:overview}. We then introduce the saliency detector and dynamic routing strategy in Sec.\,\ref{sec:saliency-detector} and Sec.\,\ref{sec:DRM}, respectively. Finally, in Sec.\,\ref{sec:loss-functions}, we detail the loss functions of the saliency detector and the SR network.


\subsection{Overview}
\label{sec:overview}
We aim to offer networks of different complexities for image patches with different saliency levels. To obtain an end-to-end network without relying on external saliency detection algorithms, we propose integrating a saliency detector into the SR framework. The unified framework makes multi-task joint learning possible. The framework of the proposed SalDRN is shown in Fig.\,\ref{fig:flowchart} and consists of three main modules: saliency detector, dynamic routing module (DRM) for feature learning, and LSUM for upsampling with any scale factor. 

Let $I_{\mathrm{LR}}$ denote the input LR image. First, the saliency detector generates a pixel-wise saliency map $I_{\mathrm{sal}}$ for $I_{\mathrm{LR}}$. Then, we decompose the input LR image $I_{\mathrm{LR}}$ and the saliency map $I_{\mathrm{sal}}$ into $N$ overlapping sub-patches of the same size and denote them as $\{P_{\mathrm{LR}}^i\}_{i=1}^N$ and $\{P_{\mathrm{sal}}^i\}_{i=1}^N$, respectively. After that, the sub-patches and their corresponding saliency maps are sent to the DRM for high-level feature extraction. The DRM contains $K$ path selection switches, each of which controls the feature map flow to or bypasses the next feature extraction block according to the average saliency value. Let $\{F_{\mathrm{DRM}}^i\}_{i=1}^N$ denote the feature maps output by the DRM, which are then recombined into a large feature map of the same size as $I_{\mathrm{LR}}$. Finally, we send the recombined high-level feature maps into LSUM to obtain the scale-arbitrary SR results $I_{\mathrm{SR}}$.

\subsection{Saliency Detector}
\label{sec:saliency-detector}

In RSIs, manufactured objects such as residential areas are very different from background areas in terms of spectral information, texture richness, and boundary shape. Saliency analysis aims to simulate the human visual perception system, which can quickly select the most attractive and discriminative areas from a large-scale scene. Therefore, we propose using saliency maps to guide the path selection in the subsequent SR network. Then, we can use varying depth networks to super-resolve regions with different visual features. This strategy can meet the different construction needs of different regions and save computing resources.

The existing algorithms proposed for the saliency detection task usually have high computational complexity due to the pursuit of high detection precision. However, our core task is saliency-aware SR, which only requires coarse region-level saliency information instead of fine saliency maps. Therefore, we designed an extremely lightweight saliency detector so that it can be integrated into the SR framework, and the additional computational burden is negligible.

\begin{figure}[t]
    \centering
    \includegraphics[width=\linewidth]{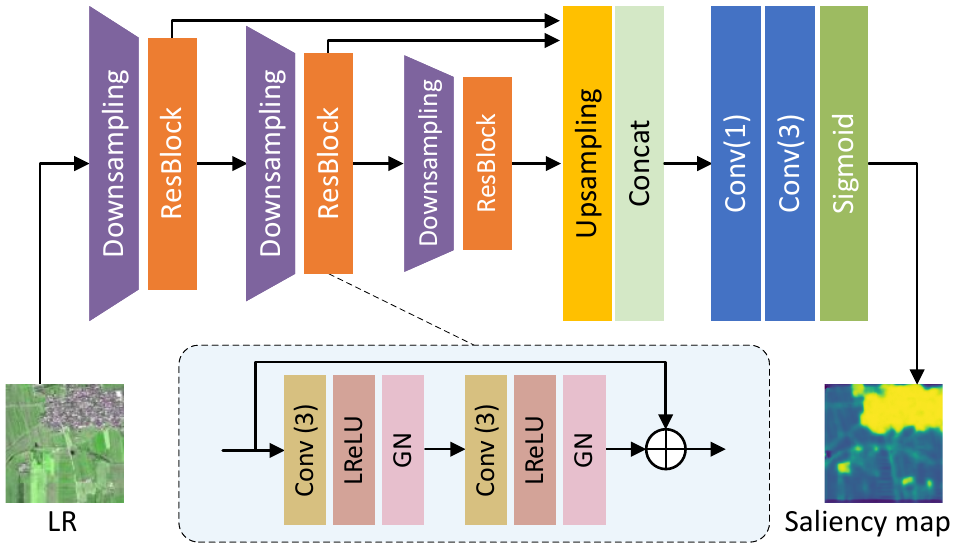}
    \caption{The architecture of the saliency detector, where ``Conv(k)'' denotes the $k\times k$ convolutional layer.}
    \label{fig:saliency_detector}
\end{figure}

The saliency detector uses an encoder-decoder architecture, as shown in Fig.\,\ref{fig:saliency_detector}. The left side of the network can be regarded as an encoder, and the right side can be regarded as a decoder. The encoder contains three residual blocks. Before each residual block, a downsampling layer is added to reduce the spatial resolution of the feature maps. The downsampling operation can effectively increase the receptive field of the shallow model and reduce the spatial dimension of the convolution operation, thus improving the computational efficiency. The downsampling layer is implemented by a $3\times 3$ convolutional layer with a stride of $2$. Specifically, the residual block is implemented with six consecutive operations and a skip connection: $\mathopr{Conv(3)}\rightarrow \mathopr{LeakyReLU}\rightarrow \mathopr{GroupNormalization}\rightarrow\mathopr{Conv(3)}\rightarrow \mathopr{LeakyReLU}\rightarrow \mathopr{GroupNormalization}$. We use group normalization \cite{wu2018group} to stabilize the training. Let $\mathopr{RB}$ denote the operation of the residual block, and let $\mathcal D$ denote the downsampling layer. Then, the outputs of the encoder $\{F_{\mathrm{enc}}^k\}_{k=1}^3$ can be calculated as
\begin{equation}
    \begin{aligned}
        F_{\mathrm{enc}}^1 & = \mathopr{RB}(\mathcal{D}(I_{LR})),                    \\
        F_{\mathrm{enc}}^k & = \mathopr{RB}(\mathcal{D}(F_{\mathrm{enc}}^{k-1})), \ \ k\in \{2,3\}. \\
    \end{aligned}
\end{equation}

Since coarse saliency maps are sufficient in our saliency-aware SR task, we design a very simple decoder to save computing resources. The decoder uses the resampling layer to adjust the size of multi-scale feature maps $\{F_k\}_{k=1}^3$ to the input image size and concatenates them. The resampling layer uses the efficient nearest interpolation. Then, we use a $1\times 1$ convolutional layer to compress the concatenated feature maps. Finally, saliency maps are obtained through a $3\times 3$ convolutional layer and a sigmoid activation function. The decoder proceeds as:
\begin{equation}
    I_{\mathrm{sal}} = \mathopr{\sigma}\Big (\mathopr{Conv}_{3,1}\big(\mathopr{Conv}_{1, m}([F_{\mathrm{enc}}^{1\uparrow}, F_{\mathrm{enc}}^{2\uparrow}, F_{\mathrm{enc}}^{3\uparrow}]\big ))\Big),
\end{equation}
where $\mathopr{Conv}_{k,m}$ denotes an $k\times k$ convolutional layer with $m$ channels, $F_{\mathrm{enc}}^{k\uparrow}$ denotes the nearest upsampling result of $F_{\mathrm{enc}}^{k}$, and $[\ \cdot\ ]$ denotes the concatenation operation. 

\subsection{Dynamic Routing Module (DRM)}
\label{sec:DRM}

The DRM aims to dynamically select the feature extraction path according to the saliency and reconstruction difficulty of image patches. Below, we introduce the dynamic routing architecture and then introduce the basic feature refinement unit (FRU).

\begin{figure}[tb]
    \centering
    \includegraphics[width=0.95\linewidth]{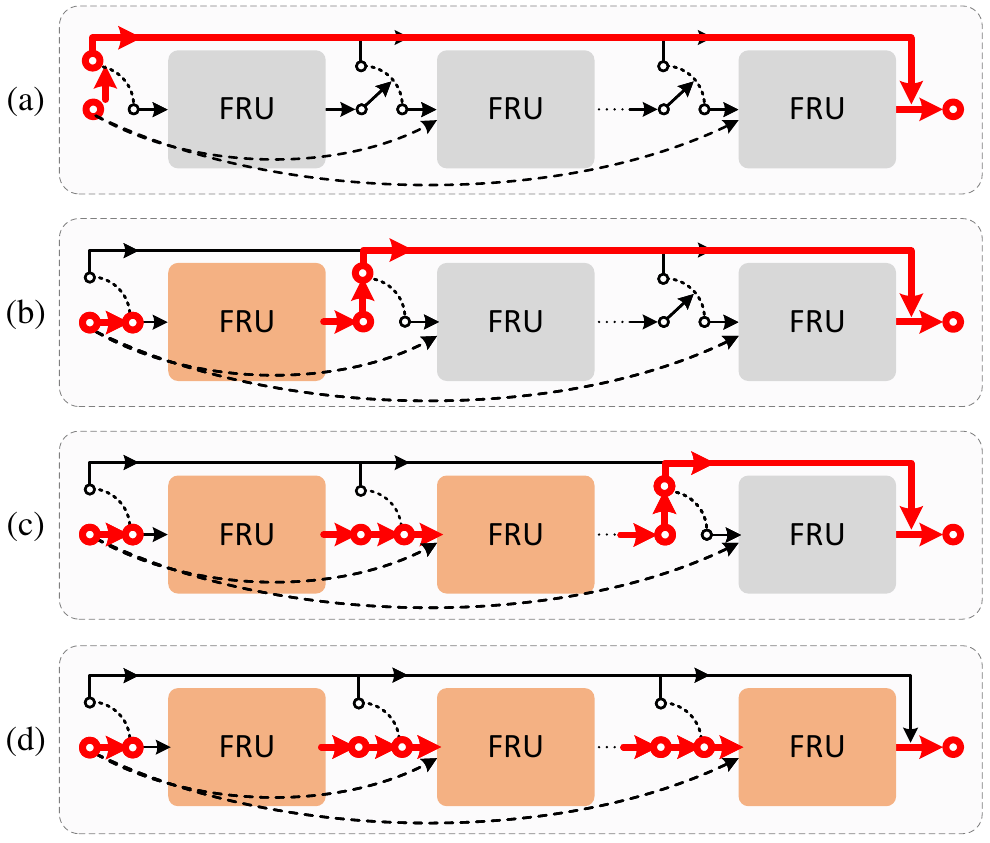}
    \caption{Different feature extraction paths of DRM.}
    \label{fig:drm}
\end{figure}

\subsubsection{Dynamic Routing Strategy} 
The DRM takes the paired LR patches and their corresponding saliency maps $\{(P_{\mathrm{LR}}^i, P_{\mathrm{sal}}^i)\}_{i=1}^N$ as inputs, and contains several stacked feature refinement units (FRU) to gradually refine the coarse low-resolution feature maps. Each FRU can directly access the original shallow feature maps and further refine the output of the previous FRU. There is a path selection switch in front of each FRU that decides whether to enter the next FRU or skip all subsequent FRUs according to the average saliency value of the image patch. In our experiments, all FRUs share the same structure and parameters. We discussed the parameter-unshared DRM in Sec.\,\ref{sec:param_sharing}. 

Specifically, assuming that the DRM contains $K$ FRUs, the passing threshold of the $k$th path selection switch is $\theta_k$. The operation after each path selection can be summarized as
\begin{equation}
    f^{(k)}(\cdot) = \Bigg\{\begin{aligned}
       & I(\cdot), \hspace{6.3em}  \text{if $\mathopr{mean}(P_{\mathrm{sal}}) \leq \theta_k$} \\
       & \mathopr{FRU}([\ \cdot\ , F_{\mathrm{shallow}}]), \hspace{0.2em} \text{else}
    \end{aligned}
\end{equation}
where $I(\cdot)$ denotes the identity mapping, and $\mathopr{FRU}(\cdot)$ represents the operation of the FRU. Suppose the sequence $\{\theta_k\}_{k=1}^K$ is monotonically increasing and there are $K+1$ possible patches in DRM, as shown in Fig.\,\ref{fig:drm}. 

\begin{figure}[tb]
    \centering
    \includegraphics[width=0.95\linewidth]{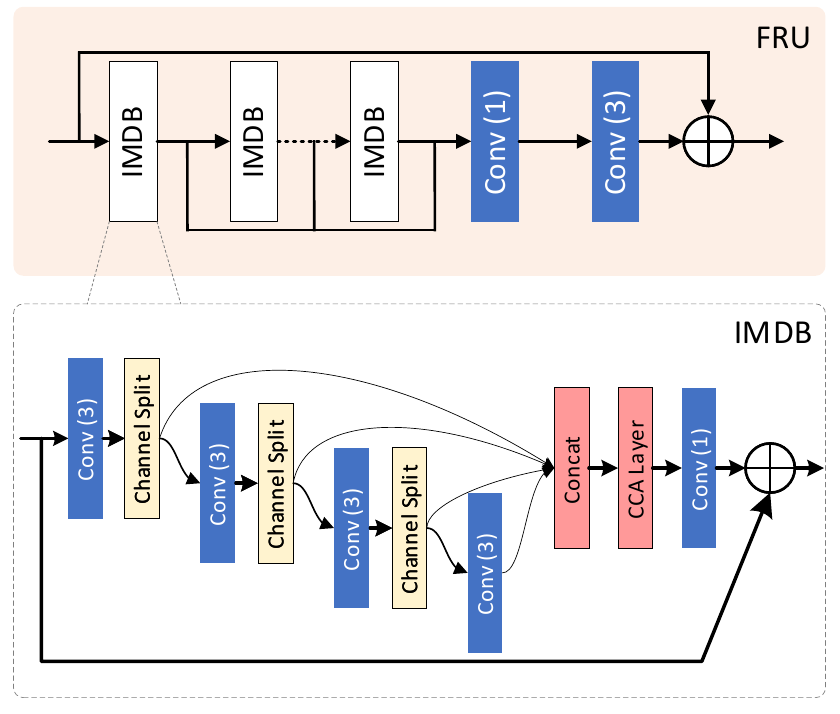}
    \caption{Illustration of the FRU.}
    \label{fig:fru}
\end{figure}

\begin{figure*}[htbp]
    \centering
    \includegraphics[width=0.98\linewidth]{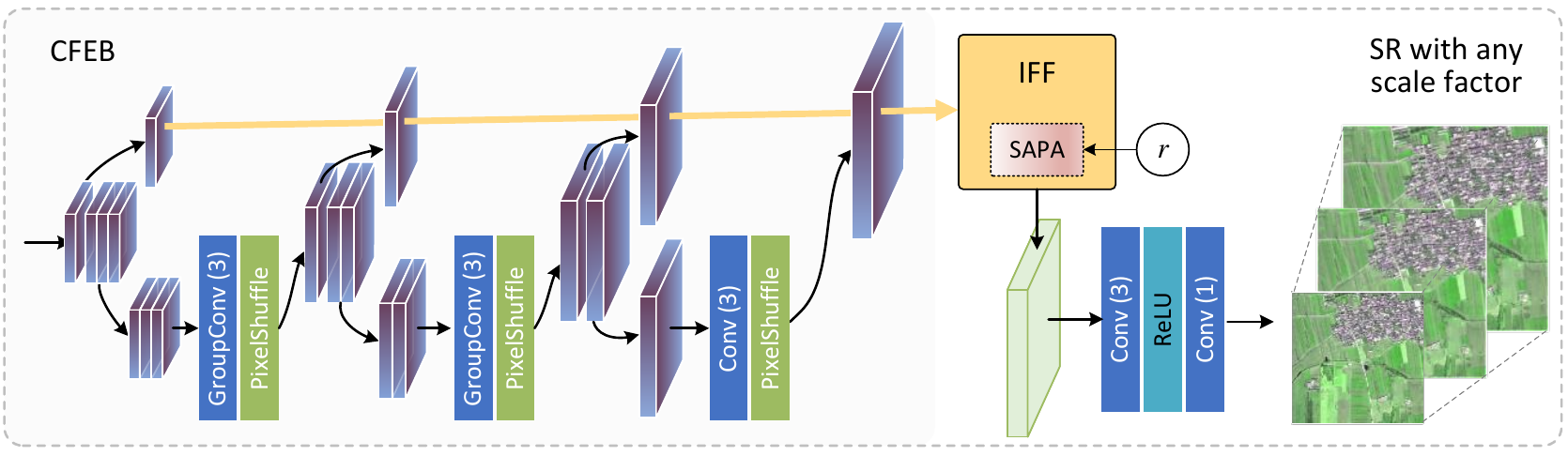}
    \caption{Illustration of the LSUM, where ``GroupConv(k)'' denotes the $k\times k$ group convolutional layer.}
    \label{fig:lsum}
\end{figure*}

\subsubsection{FRU} The architecture of the FRU is shown in Fig.\,\ref{fig:fru}. We choose the information multi-distillation block (IMDB) \cite{hui2019lightweight} as the basic feature extraction block due to its superiority in previous lightweight SR tasks. The FRU contains a global residual connection from start to end and contains $D$ multiple stacked IMDBs. The input feature maps are sequentially refined by IMDBs, and then all intermediate outputs are integrated using a $1\times 1$ convolutional layer followed by a $3\times 3$ convolutional layer. The procedure can be summarized as follows:
\begin{equation}
    \begin{aligned}
        H_d              & = \mathopr{IMDB}_d(H_{d-1}),\ \ d = 1,\cdots, D                                    \\
        H_{\mathrm{out}} & = \mathopr{Conv}_{3,C}\big(\mathopr{Conv}_{1,C}([H_1,H_2,\cdots, H_D])\big) + H_0, \\
    \end{aligned}
\end{equation}
where $H_0$ and $H_{\mathrm{out}}$ represent the input and output of FRU, respectively. $\mathopr{IMDB}(\cdot)$ denotes the operation of IMDB. 

In each IMDB, we progressively refine high-level features using four convolutional layers, as shown in the lower part of Fig.\,\ref{fig:fru}. After each convolutional layer (except the last layer), a channel-splitting operation decomposes the feature maps into two parts. One part of the refined features is kept, and the other part is processed in the next convolutional layer. Let $G_{\mathrm{in}}$ denote the input of IMDB. Then, the $d$th operation of IMDB can be calculated as
\begin{equation}
    \begin{aligned}
        G_{\mathrm{refined}\_1}^{d}, G_{\mathrm{coarse}\_1}^{d} & = \mathopr{Split}(\mathopr{Conv}_{3,C}(G_{\mathrm{in}})),           \\
        G_{\mathrm{refined}\_2}^{d}, G_{\mathrm{coarse}\_2}^{d} & = \mathopr{Split}(\mathopr{Conv}_{3,C}(G_{\mathrm{coarse}\_1}^{d}), \\
        G_{\mathrm{refined}\_3}^{d}, G_{\mathrm{coarse}\_3}^{d} & = \mathopr{Split}(\mathopr{Conv}_{3,C}(G_{\mathrm{coarse}\_2}^{d}), \\
        G_{\mathrm{refined}\_4}^{d}                             & = \mathopr{Conv}_{3,C/4}(G_{\mathrm{coarse}\_3}^{d}),               \\
    \end{aligned}
\end{equation}
where $\mathopr{Split}(\cdot)$ represents the channel-splitting operation, and $G_{\mathrm{refined}\_i}^{d}$ represents the $i$th refined feature maps with channel number $C/4$. Each convolution operation $\mathopr{Conv}_{3,C}(\cdot)$ contains a leaky rectified linear unit (LReLU) as the activation function. Then, the four refined feature maps are concatenated:
\begin{equation}
    G_{\mathrm{distilled}}^d = [G_{\mathrm{refined}\_1}^{d}, G_{\mathrm{refined}\_2}^{d}, G_{\mathrm{refined}\_3}^{d}, G_{\mathrm{refined}\_4}^{d}],
\end{equation}
and the result $G_{\mathrm{distilled}}^d$ is fed into the contrast-aware channel attention (CCA) layer. Finally, a $1\times 1$ convolutional layer is adopted to facilitate the flow of information across channels, and the result is then added to the original input feature to obtain the output of IMDB. 

The CCA layer is an improvement of the squeezing and excitation module (SE-Block) \cite{hu2018squeeze}, replacing the global average pooling with a global contrast information function. This preserves more low-level visual information such as structure, texture, and edges for SR tasks.

Let $x_c\in\mathbb R^{1\times H\times W}$ denote the $c$th channel of the feature maps $X\in\mathbb R^{C\times H\times W}$, where $(H,W)$ represents the height and width of the image. The global contrast information function $f_{\mathrm{GC}}$ is defined as follows:
\begin{equation}
    \label{eq:L3X2}
    \begin{aligned}
        z_c = & f_{\mathrm{GC}}(x_c)                                                                                                                                 \\
             = & \sqrt{\frac{1}{HW}\sum_{(i,j)\in x_c}\Bigg(x_c^{i,j} - \frac{1}{HW}\sum_{(i,j)\in x_c}x_c^{i,j}\Bigg)^2} \\
            & +  \frac{1}{HW}\sum_{(i,j)\in x_c}x_c^{i,j}.
    \end{aligned}
\end{equation}
We compress the input feature maps $X$ into a vector $Z=[z_1, z_2, \cdots, z_C]\in\mathbb R^{C\times 1\times 1}$ using \eqref{eq:L3X2}. Then, a three-layer MLP is adopted to obtain channel-wise attention weights. The numbers of neurons in the hidden layer and output layer are $C/16$ and $C$, respectively. Finally, attention weights are normalized to $[0,1]$ using a sigmoid function. The input feature maps $X$ are reweighted using the channel attention weights, thus emphasizing features that are more beneficial for the SR task.

\subsection{ Lightweight Stepless Upsampling Module (LSUM)}
\label{sec:lsum}

To cope with potential decimal-scale factors, a simple idea is to use interpolation algorithms such as bilinear or bicubic to resize the feature maps. However, upsampling the input LR image to the target size in advance will significantly increase the computational complexity. Additionally, Hu \etal \cite{hu2019meta} proved that interpolating the feature maps at the end of the network will provide poor results. This problem occurs mainly because the interpolation algorithms cannot effectively map the features of a single scale to other scales. To address this problem, we propose constructing a family of feature maps with increasing resolution, and designing implicit feature functions (IFFs) to obtain a continuous representation of the image in the high-level feature space.

Fig.\,\ref{fig:lsum} presents an overview of the LSUM. First, we design a cascade frequency enhancement block (CFEB) to obtain progressive resolution feature maps $\{M_t \in \mathbb R^{\frac{C}{4}\times tH \times tW}\}_{t=1}^4$. CFEB uses group convolutions and channel split operations to reduce computational complexity. Then, we introduce IFFs to map any 2D coordinate to a vector in the continuous feature space. We also propose a novel scale-aware pixel-level attention (SAPA) mechanism for IFF and combine it with scale factor encoding, enabling efficient feature mapping across scales. Thus, the continuous feature representation of the image is obtained with a high-resolution feature map $F_{\mathrm{HR}}^{(r)}$ for any scale factor of $r$. Finally, we obtain the final SR result by using two consecutive convolutional layers:
\begin{equation}
    I_{\mathrm{SR}} = \mathopr{Conv}_{1,3}(\mathopr{ReLU}(\mathopr{Conv}_{3,C/4}(F_{\mathrm{HR}}^{(r)}))).
\end{equation}


\subsubsection{CFEB} CFEB aims to obtain a set of feature maps with progressive resolution by gradually adding high-frequency details, as shown in the left part of Fig.\,\ref{fig:lsum}. 

First, we split the input feature maps $M_{\mathrm{in}} \in \mathbb{R}^{C\times H \times W}$ into two parts along the channel dimension, one of which contains $1/4$ of the feature maps, denoted as $M_1 \in \mathbb{R}^{\frac {C}{4} \times H \times W}$, and kept as the current resolution. We use group convolution and pixel-shuffling to add high-frequency details to the remaining $3/4$ feature maps $\tilde{M}_1 \in \mathbb{R}^{\frac {3C}{4} \times H \times W}$ and increase the spatial resolution by a scale factor of $2$. Then, we split the feature maps of the second-level resolution into two parts and keep $1/3$ of the feature maps $M_2 \in \mathbb{R}^{\frac{C}{4} \times 2H \times 2W}$ as the current resolution. The resolution of the remaining $2/3$ feature maps $\tilde{M}_2 \in \mathbb{R}^{\frac{2C}{4} \times 2H \times 2W}$ is again doubled by using group convolution and pixel-shuffling. Finally, the third-level resolution feature maps are split into two halves along the channel dimension. Half of the feature maps $M_3 \in \mathbb{R}^{\frac{C}{4} \times 4H \times 4W}$ are kept at the current resolution, and the other half $\tilde{M}_3 \in \mathbb{R}^{\frac{C}{4} \times 4H \times 4W}$ are upscaled twice using a convolutional layer followed by a pixel-shuffle operation. The last-level resolution is $8$ times the original input, and the last-level feature maps are denoted as $M_4 \in \mathbb{R}^{\frac{C}{4} \times 8H \times 8W}$. Thus, we obtain feature maps $\{M_t\}_{t=1}^4$ with progressive resolution. The calculation of the CFEB can be summarized as \eqref{eq:FOoY}
\begin{equation}
    \label{eq:FOoY}
    \begin{aligned}
        M_1, \tilde{M}_1 & = \mathopr{Split}(M_{\mathrm{in}}),                                                      \\
        M_2, \tilde{M}_2 & = \mathopr{Split}(\mathopr{PixShuffle}(\mathopr{GroupConv}_{3,3C}^{(3)}(\tilde{M}_1)), \\
        M_3, \tilde{M}_3 & = \mathopr{Split}(\mathopr{PixShuffle}(\mathopr{GroupConv}_{3,2C}^{(2)}(\tilde{M}_2)), \\
        M_4              & =\mathopr{PixShuffle}(\mathopr{Conv}_{3,C}(\tilde{M}_3)),
    \end{aligned}
\end{equation}
where $\mathopr{GroupConv}_{k,C}^{(n)}(\cdot)$ denotes a group convolutional layer with a kernel size of $k\times k$, number of channels $C$, and number of groups $n$. $\mathopr{PixShuffle}(\cdot)$ denotes the pixel-shuffle operation.

\subsubsection{IFF}

Implicit neural representation is a continuous and differentiable function parameterized by neural networks, which has been widely used in 3D tasks \cite{chen2019learning,sitzmann2019scene,mildenhall2020nerf}. Our work is inspired by LIIF \cite{chen2021learning}, which learns an implicit function of an image that maps any coordinate to a pixel value, enabling SR with arbitrary scale factors. Since the implicit function is shared for all images, it takes the feature vectors of pixels near the query coordinate, called latent codes, as additional input. LIIF directly uses implicit functions to predict pixel values of the HR image, resulting in a complex network structure of implicit function and high computational cost. In addition, since the implicit functions in LIIF are constructed using only one latent code of a single scale, a local ensemble strategy is required to avoid discontinuous artifacts, which further increases the complexity of the upsampling module. 

To address the above problems, we propose improvements in three aspects. First, we let the implicit function map from 2D coordinates to HR feature vectors instead of pixel values, which can greatly reduce the difficulty of learning the mapping. Second, we construct the implicit functions using a local set of latent codes from multi-resolution spaces, which can avoid the artifact problem at a low computational cost. Third, we introduce a scale-aware pixel attention mechanism incorporated with periodic scale encoding, which can effectively improve the performance of multi-scale factor tasks. 

\begin{figure}[t]
    \centering
    \includegraphics[width=0.65\linewidth]{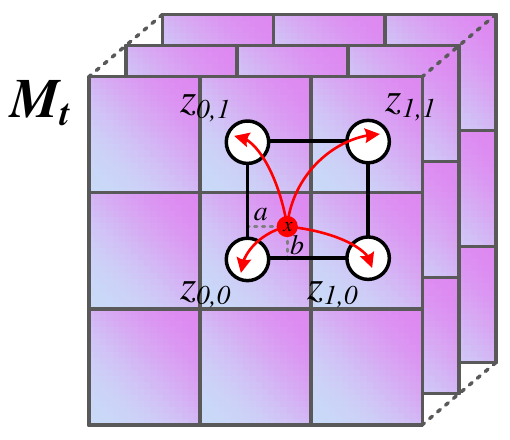}
    \caption{Illustration of latent codes of a query coordinate $x$ in $M_t$.}
    \label{fig:latentcode}
\end{figure}

Formally, let $f_{\mathrm{IFF}}$ denote the IFF, which maps any 2D coordinate $x$ to the feature vector $\mathbf{y}_{x}$ of the corresponding position. Recall that we obtained progressive resolution feature maps $\{M_1, \cdots, M_4\}$ from CFEB. Then the continuous feature representation is defined as
\begin{equation}
    \begin{aligned}
        \mathbf{y}_{x} & = f_{\mathrm{IFF}}(x; M_1, M_2, M_3, M_4)                                                                   \\
                       & = \mathopr{SAPA}(r, [h(x, \mathcal{Z}_1), h(x, \mathcal{Z}_2),  h(x, \mathcal{Z}_3), h(x, \mathcal{Z}_4)]),
    \end{aligned}
\end{equation}
where $\mathopr{SAPA}(r, \cdot)$ denotes a scale-aware pixel attention layer with a scale factor of $r$, and $\mathcal{Z}_t=\{z_{0,0}, z_{0,1}, z_{1,0}, z_{1,1}\}$ is a set of four nearest (Euclidean distance) latent codes from $x$ in $M_t$, i.e., the feature vectors of the four pixels nearest to the query coordinate $x$. Fig.\,\ref{fig:latentcode} illustrates the relationship between a query coordinate and latent codes. Let $z_{0,0}, z_{0,1}, z_{1,0}, z_{1,1} \in \mathbb{R}^{\frac{C}{4}\times 1\times 1}$ ($t$ and $x$ are omitted) denote the lower left, upper left, lower right, and upper right latent codes of the query coordinate $x$, respectively. $h(x, \mathcal{Z}_t)$ is a local bilinear function that interpolates the four latent codes in both horizontal and vertical dimensions and is defined as follows:
\begin{equation}
    h(x, \mathcal Z_t) =
    \begin{bmatrix}
        1-a & a
    \end{bmatrix}
    \begin{bmatrix}
        z_{0,0} & z_{0,1} \\
        z_{1,0} & z_{1,1}
    \end{bmatrix}
    \begin{bmatrix}
        1-b \\
        b
    \end{bmatrix},
\end{equation}
where $(a,b)$ represents the coordinates of $x$ relative to $z_{0,0}$ normalized to $[0, 1]$. 

\subsubsection{SAPA} Simply concatenating the feature vectors $\{h(x, \mathcal Z_t)\}_{t=1}^4$ from spaces of different resolutions will not provide an ideal solution because the importance of input feature vectors varies under different scale factors. It is easy to see that when the scale factor is relatively large, such as $r=4$, $h(x,\mathcal Z_3)$ and $h(x,\mathcal Z_4)$ should be assigned higher attention weights; otherwise, $h(x,\mathcal Z_1)$ and $h(x,\mathcal Z_2)$ should be assigned higher attention weights. Therefore, to adaptively adjust the attention weights of the concatenated feature vectors according to $r$, thereby enhancing the representation ability of IFF in multi-task learning, we introduce scale-aware pixel attention. 

Recent studies \cite{mildenhall2020nerf,xu2021ultrasr} showed that low-level information encoding is one of the keys to generating high-frequency details in 2D/3D tasks.  In our scale arbitrary SR task, the scale factor $r$ is critical input information that will affect the distribution of HR feature spaces. Therefore, we employ periodic encoding to expand 1D $r$ into a 64D vector,
\begin{equation}
    \begin{aligned}
        \phi(r) = [\sin(\omega_1 r), \cos(\omega_1 r), &\sin(\omega_2 r), \cos(\omega_2 r), \\
         \cdots,  &\sin(\omega_{32} r), \cos(\omega_{32} r)],    
    \end{aligned}
\end{equation}
where $\omega_1,\omega_2,\cdots\omega_{32}$ are frequency parameters, and $\omega_i = 2e^{i}$. 

Then, the feature vectors $\{h(x, \mathcal Z_t)\}_{t=1}^4$ are concatenated with the scale encoding $\phi(r)$ and fed to an MLP with a single hidden layer to calculate the pixel-level attention weights. The number of hidden layer neurons in the MLP is $C/2$, and LReLU is adopted as the activation function. The attention weights $\bm\alpha$ are calculated as follows:
\begin{equation}
    \bm \alpha = \mathopr{MLP} ([\phi(r), h(x, \mathcal{Z}_1), h(x, \mathcal{Z}_2),  h(x, \mathcal{Z}_3), h(x, \mathcal{Z}_4)]),
\end{equation}
where $\mathopr{MLP}(\cdot)$ represents the operation of the three-layer MLP. Finally, we multiply the attention weights with the input feature vector to dynamically emphasize information at different scales. The output of the IFF $\mathbf{y}_{x}$ can be calculated as
\begin{equation}
    \mathbf{y}_{x} = [h(x, \mathcal{Z}_1), h(x, \mathcal{Z}_2),  h(x, \mathcal{Z}_3), h(x, \mathcal{Z}_4)] \otimes \bm \alpha,
\end{equation}
where $\otimes$ is the element-wise multiplication. 

\subsection{Loss Functions}
\label{sec:loss-functions}
hl{here}
The loss function of the SalDRN consists of three parts: SR loss $\mathcal L_{\mathrm{SR}}$, saliency loss $\mathcal L_{\mathrm{sal}}$, and reconstruction difficulty loss $\mathcal L_{\mathrm{diff}}$. The SR loss guides the network to generate content-correct SR results, while the saliency loss enables the saliency detector to generate accurate saliency maps and guides path selection in the DRM. The reconstruction difficulty loss can provide the saliency detector with the reconstruction difficulty information of the image patches according to the error of the SR result. The total loss of the network is defined as
\begin{equation}
    \mathcal{L} = \mathcal{L}_{\mathrm{SR}} + \lambda_1\mathcal{L}_{\mathrm{sal}} +\lambda_2 \mathcal{L}_{\mathrm{diff}},
\end{equation}
where $\lambda_1$ and $\lambda_2$ are hyperparameters to balance the three sub-items.

\subsubsection{SR Loss}
DRM has $K+1$ possible path choices, and the $k$th path contains $k-1$ FRUs. In the testing phase, each image patch selects the most suitable feature extraction path according to the average saliency value. However, in the training phase, we pass the LR image patch $P_{\mathrm{LR}}$ through all possible paths so that the parameters of all paths can be optimized simultaneously. Therefore, we obtain $K+1$ SR results $\{P_{\mathrm{SR}}^k\}_{k=1}^{K+1}$ and then compute the $L_1$ losses by comparing these SR results with the HR image patch $P_{\mathrm{HR}}$. The SR loss is defined as
\begin{equation}
    \label{eq:b2Ae}
    \mathcal{L}_{\mathrm{SR}} = \sum_{k=1}^{K+1}\beta_k(s)\|P_{\mathrm{SR}}^k - P_{\mathrm{ HR}}\|_1,
\end{equation}
where $\|\cdot\|_1$ denotes the $L_1$ norm, and $\beta_k$ is the weight of the $k$th path loss function, which is defined according to the probability of the path being selected and will be described in detail below. 

Let $s$ denote the mean saliency value of an image patch, i.e.,
$
    s = \mathopr{mean}(P_{\mathrm{sal}}).
$
The larger the saliency value $s$, the higher the reconstruction difficulty of the image patch, so the path containing more FRUs should be selected with a greater probability. Recall that in DRM, $\theta_k$ denotes the threshold of the $k$ path selection switch. 
We define $\theta_{K+1} = 1$. Then, the weights in \eqref{eq:b2Ae} can be calculated as follows:
\begin{equation}
    \label{eq:bhYR}
    \begin{aligned}
        \tilde\beta_k(s) & = \gamma (\theta_{k+1} - s)(s - \theta_{k}), \ \ k=1,2,\cdots,K+1 \\
        \beta_k(s) & = \frac{\exp(\tilde\beta_k)}{\sum_{i=1}^{K+1}\exp(\tilde\beta_i)}, \ \ k=1,2 ,\cdots,K+1
    \end{aligned}
\end{equation}
where $\gamma$ is a hyperparameter that adjusts the difference in weights. We empirically set $\gamma=10$ to obtain satisfactory results. It can be seen from \eqref{eq:bhYR} that $\beta_k$ takes the maximum if and only if $\theta_k \leq s \leq \theta_{k+1}$. This property guarantees that the definition of $\beta_k$ is reasonable, and the path within the threshold range can be selected with the greatest probability.

\subsubsection{Saliency Loss} 
The saliency loss $\mathcal L_{\mathrm{sal}}$ is calculated by the cross-entropy loss between the predicted saliency maps and the saliency ground truth (GT):
\begin{equation}
    \begin{aligned}
        \mathcal{L}_{\mathrm{sal}} = -\frac{1}{|\mathcal P|} \sum_{i=1}^{|\mathcal P|} 
        \big(L_{\mathrm{sal}}^{i}\log I_{\mathrm{sal}}^{i}
         + (1-L_{\mathrm{sal}}^{i})\log (1-I_{\mathrm{sal}}^{i})\big),
    \end{aligned}
\end{equation}
where $L_{\mathrm{sal}}^{i}$ and $I_{\mathrm{sal}}^{i}$ represent the saliency GT of the $i$th pixel and the predicted saliency value, respectively, and $|\mathcal P|$ is the number of pixels in the training image. 

\subsubsection{Difficulty Loss}
We also use the reconstruction error as a guide for saliency detection. Regions with large differences between the SR results and the GT indicate that reconstruction is difficult, and higher saliency values should be predicted to guide image patches to select deeper feature extraction paths. The error map $I_{\mathrm{err}}$ is calculated as follows: first, calculate the mean square error between the SR result and the GT; then, apply mean filtering to the error map; finally, histogram equalization is performed on the error map so that the error distribution histogram is approximately uniform. The reconstruction difficulty loss is defined as the $L_1$ loss of the predicted saliency map and error map,
\begin{equation}
     \mathcal{L}_{\mathrm{diff}} = \|I_{\mathrm{sal}} - I_{\mathrm{err}}\|_1.
\end{equation}
\section{Experimental Results}
\label{sec:experiments}

In this section, we introduce the datasets, metrics, and training details. Then, we compare our proposal with SOTA fixed-scale/stepless SR models. Finally, we compare and analyze the computational complexity of our proposal and competitive models. 

\subsection{Datasets and Metrics}

We use remote sensing imagery provided by the GeoEye-1 satellite and Google Earth to verify the effectiveness of the proposed method. The GeoEye-1 dataset contains $130$ multispectral images with a resolution of $\SI{0.41}{\metre}$ and a size of $512\times 512$, of which $115$ images are used for training, and $15$ images constitute a test set. The Google Earth dataset contains $239$ optical RSIs with a resolution of $\SI{1}{\metre}$ and a size of $512\times 512$, of which $224$ are used for training and $15$ images constitute a test set. In the experiments of this study, the training set contains a total of $339$ RSIs from the above two sources.

\subsection{Training Details}

Our proposed model is implemented based on the PyTorch framework and trained on an NVIDIA GeForce RTX 3090 GPU. 

In the testing phase, we crop the input image into patches with a size of $48\times 48$ and an overlap of $8$ pixels to enforce the dynamic routing strategy. The basic number of channels $C$ in the feature extraction part is set to $64$. The number of FRUs $K$ is set to $3$, and the thresholds of the path selection switches are set to $0, 0.25, 0.5$. The number of IMDBs $D=4$ in each FRU, and the parameters are shared between FRUs. The weight parameters $\lambda_1, \lambda_2$ in the loss function are set to $0.1$ and $0.15$, respectively.

In the training phase, LR images are obtained using bicubic downsampling and randomly cropped into $32\times 32$ image patches as input. We randomly flip input patches vertically, horizontally, or rotate $90^{\circ}$ for data augmentation. We adopt Adam \cite{kingma2014adam} as the parameter optimizer, setting $\beta_1 = 0.1$. The model is trained with a mini-batch size of $16$, and each batch of LR images has the same scale factor. The scale factor during training follows a uniform distribution of $[1, 4]$. We train our model for $4\times 10^5$ iterations with an initial learning rate of $1\times 10^{-4}$, and the learning rate is halved at the $2\times 10^5$ iteration. 

\begin{table}[tbp]
    \centering
    \renewcommand\arraystretch{1.2}
    \caption{PSNR(\si{\dB}) and SSIM comparison of integer scale SR results. The Best Performance is Shown in \textbf{Bold} and the Second Best \underline{Underlined}.}
    \label{tab:results-fs}
    \setlength{\tabcolsep}{1.4mm}{
        \begin{tabular}{l|c|c|cc|cc}
            \hline
            \hline
            \multicolumn{1}{l|}{\multirow{2}[1]{*}{Methods}} & \multirow{2}[1]{*}{Scale} & \multirow{2}[1]{*}{Params} & \multicolumn{2}{c|}{GeoEye-1} & \multicolumn{2}{c}{Google Earth} \bigstrut[t]\\
            &    &    & PSNR & SSIM & PSNR & SSIM \bigstrut[b]\\
            \hline
            Bicubic & $\times 2$ & -  & 24.84  & 0.7900  & 28.03  & 0.8243  \bigstrut[t]\\
            CARN \cite{ahn2018fast} & $\times 2$ & 1592K & 26.69  & 0.8583  & \underline{30.00}  & 0.8818  \\
            IMDN \cite{hui2019lightweight} & $\times 2$ & 694K & 26.70  & 0.8573  & 29.98  & 0.8816  \\
            CTN \cite{wang2021contextual} & $\times 2$ & 412K & 26.67  & 0.8585  & 29.80  & 0.8782  \\
            LatticeNet \cite{luo2020latticenet} & $\times 2$ & 765K & 26.67  & 0.8579  & 29.94  & 0.8807  \\
            PAN \cite{zhao2020efficient} & $\times 2$ & 272K & 26.66  & 0.8581  & 29.89  & 0.8798  \\
            CFSRCNN \cite{tian2020coarse} & $\times 2$ & 1310K & 26.72  & 0.8594  & \underline{30.00}  & \textbf{0.8823} \\
            FeNet \cite{wang2022fenet} & $\times 2$ & 618K & 26.73  & 0.8594  & 29.95  & 0.8808 \\
            SalDRN (Ours) & $\times 2$ & 571K & \textbf{26.80}  & \textbf{0.8617} & 29.97  & 0.8811  \\
            SalDRN-FS (Ours) & $\times 2$ & 519K & \underline{26.79}  & \underline{0.8604}  & \textbf{30.01} & \underline{0.8818} \bigstrut[b] \\
            \hline
            Bicubic & $\times 3$ & -  & 22.52  & 0.6414  & 25.25  & 0.6841  \bigstrut[t]\\
            CARN \cite{ahn2018fast} & $\times 3$ & 1592K & 23.93  & 0.7404  & 26.96  & 0.7710  \\
            IMDN \cite{hui2019lightweight} & $\times 3$ & 703K & 23.97  & 0.7422  & 26.99  & 0.7719  \\
            CTN \cite{wang2021contextual} & $\times 3$ & 412K & 23.83  & 0.7365  & 26.83  & 0.7679  \\
            LatticeNet \cite{luo2020latticenet} & $\times 3$ & 765K & 23.95  & 0.7414  & 26.93  & 0.7706  \\
            PAN \cite{zhao2020efficient} & $\times 3$ & 272K & 23.88  & 0.7377  & 26.86  & 0.7679  \\
            CFSRCNN \cite{tian2020coarse} & $\times 3$ & 1495K & \underline{23.98}  & 0.7429  & \underline{27.01}  & \underline{0.7729}  \\
            FeNet \cite{wang2022fenet} & $\times 3$ & 627K & 23.91  & 0.7407  & 26.94  & 0.7705 \\
            SalDRN (Ours) & $\times 3$ & 571K & \underline{23.98}  & \underline{0.7437 } & 26.97  & 0.7722  \\
            SalDRN-FS (Ours) & $\times 3$ & 519K & \textbf{24.02} & \textbf{0.7438} & \textbf{27.03} & \textbf{0.7734} \bigstrut[b]\\
            \hline
            Bicubic & $\times 4$ & -  & 21.25  & 0.5307  & 23.74  & 0.5800  \bigstrut[t]\\
            CARN \cite{ahn2018fast} & $\times 4$ & 1592K & 22.39  & 0.6383  & 25.15  & 0.6761  \\
            IMDN \cite{hui2019lightweight} & $\times 4$ & 715K & 22.43  & 0.6408  & 25.19  & 0.6780  \\
            CTN \cite{wang2021contextual} & $\times 4$ & 412K & 22.30  & 0.6324  & 25.04  & 0.6698  \\
            LatticeNet \cite{luo2020latticenet} & $\times 4$ & 765K & 22.44  & 0.6418  & 25.16  & 0.6765  \\
            PAN \cite{zhao2020efficient} & $\times 4$ & 272K & 22.36  & 0.6373  & 25.11  & 0.6739  \\
            CFSRCNN \cite{tian2020coarse} & $\times 4$ & 1458K & \underline{22.49}  & \textbf{0.6453} & \underline{25.22}  & \textbf{0.6799} \\
            FeNet \cite{wang2022fenet} & $\times 4$ & 639K & 22.39  & 0.6371  & 25.16  & 0.6753 \\
            SalDRN (Ours) & $\times 4$ & 571K & 22.42  & 0.6408  & 25.17  & 0.6768 \\
            SalDRN-FS (Ours) & $\times 4$ & 530K & \textbf{22.51} & \underline{0.6449}  & \textbf{25.23} & \underline{0.6798}  \bigstrut[b]\\
            \hline
            \hline
        \end{tabular}%
    }
\end{table}

\begin{table*}
    \centering
    \renewcommand\arraystretch{1.2}
    \caption{PSNR(\si{\dB}) comparison of continuous-scale SR results. The Best Performance is Shown in \textbf{Bold} and the Second Best \underline{Underlined}.}
    \setlength{\tabcolsep}{1.8mm}{
    \label{tab:results-ms}
    \begin{tabular}{cl|cccccccccccc}
    \hline
    \hline
    Datasets & Methods & $\times$1.1 & $\times$1.2 & $\times$1.4 & $\times$1.5 & $\times$2.3 & $\times$2.4 & $\times$2.6 & $\times$2.8 & $\times$3.1 & $\times$3.6 & $\times$3.7 & $\times$3.9 \bigstrut\\
    \hline
    \multirow{9}[2]{*}{Google Earth} & Bicubic & 34.97  & 33.44  & 31.45  & 30.68  & 27.00  & 26.71  & 26.14  & 25.67  & 25.08  & 24.26  & 24.12  & 23.85  \bigstrut[t]\\
       & CARN\cite{ahn2018fast} + LIIF\cite{chen2021learning} & 38.76  & 36.08  & 33.48  & 32.61  & 28.77  & 28.47  & 27.87  & 27.36  & 26.68  & 25.74  & 25.56  & 25.24  \\
       & CARN\cite{ahn2018fast} + OSM\cite{behjati2021overnet} & 38.71  & 36.01  & 33.46  & 32.60  & 28.77  & 28.45  & 27.87  & 27.36  & 26.67  & 25.72  & 25.56  & 25.24  \\
       & CARN\cite{ahn2018fast} + LSUM (Ours) & 38.87  & 36.09  & 33.53  & 32.65  & 28.78  & 28.45  & 27.87  & 27.36  & 26.68  & 25.72  & 25.55  & 25.24  \\
       & IMDN\cite{hui2019lightweight} + LIIF\cite{chen2021learning} & 38.74  & 36.09  & 33.50  & 32.63  & 28.81  & 28.50  & 27.91  & 27.39  & 26.72  & 25.77  & \underline{25.59}  & \underline{25.28}  \\
       & IMDN\cite{hui2019lightweight} + OSM\cite{behjati2021overnet} & 38.90  & 36.00  & 33.54  & 32.67  & \underline{28.83}  & \underline{28.51}  & \underline{27.93}  & \underline{27.42}  & \underline{26.73}  & \underline{25.77}  & \underline{25.59}  & 25.27  \\
       & IMDN\cite{hui2019lightweight} + LSUM (Ours) & \textbf{38.98}  & \underline{36.14}  & \textbf{33.57}  & \underline{32.68}  & 28.81  & 28.48  & 27.90  & 27.39  & 26.70  & 25.76  & \underline{25.59}  & 25.27  \\
       & OverNet\cite{behjati2021overnet} & 38.91  & 36.11  & 33.54  & 32.66  & 28.78  & 28.45  & 27.87  & 27.36  & 26.67  & 25.73  & 25.54  & 25.24  \\
       & SalDRN (Ours) & \underline{38.97}  & \textbf{36.15}  & \textbf{33.57}  & \textbf{32.70}  & \textbf{28.87}  & \textbf{28.55}  & \textbf{27.97}  & \textbf{27.45}  & \textbf{26.78}  & \textbf{25.80}  & \textbf{25.62}  & \textbf{25.30}  \bigstrut[b]\\
    \hline
    \multirow{9}[2]{*}{GoeEye-1} & Bicubic & 31.08  & 28.61  & 26.58  & 25.24  & 24.26  & 23.28  & 23.07  & 22.70  & 22.23  & 21.93  & 21.80  & 21.44  \bigstrut[t]\\
       & CARN\cite{ahn2018fast} + LIIF\cite{chen2021learning} & 35.45  & 31.39  & 28.73  & 27.11  & 25.92  & 24.74  & 24.52  & 24.08  & 23.50  & 23.15  & 22.99  & 22.53  \\
       & CARN\cite{ahn2018fast} + OSM\cite{behjati2021overnet} & 34.99  & 31.02  & 28.32  & 26.76  & 25.66  & 24.61  & 24.39  & 23.99  & 23.45  & 23.10  & 22.96  & 22.49  \\
       & CARN\cite{ahn2018fast} + LSUM (Ours) & 35.52  & 31.39  & 28.69  & 27.05  & 25.85  & 24.67  & 24.43  & 23.99  & 23.42  & 23.08  & 22.93  & 22.47  \\
       & IMDN\cite{hui2019lightweight} + LIIF\cite{chen2021learning} & 35.49  & 31.47  & 28.81  & 27.18  & 26.00  & 24.81  & \underline{24.59}  & \underline{24.14}  & 23.54  & \underline{23.18}  & \underline{23.03}  & \textbf{22.61}  \\
       & IMDN\cite{hui2019lightweight} + OSM\cite{behjati2021overnet} & 33.95  & 31.41  & 28.81  & 27.17  & 25.99  & 24.82  & \underline{24.59}  & 24.13  & 23.53  & 23.16  & 23.02  & 22.57  \\
       & IMDN\cite{hui2019lightweight} + LSUM (Ours) & \underline{35.76}  & \underline{31.53}  & \underline{28.83}  & \underline{27.20}  & \underline{26.02}  & \underline{24.83}  & \underline{24.59}  & \underline{24.14}  & \underline{23.55}  & 23.17  & 23.02  & 22.57  \\
       & OverNet\cite{behjati2021overnet} & 33.66  & 31.38  & 28.79  & 27.14  & 25.94  & 24.75  & 24.52  & 24.08  & 23.48  & 23.11  & 22.96  & 22.52  \\
       & SalDRN (Ours) & \textbf{35.91}  & \textbf{31.63}  & \textbf{28.89}  & \textbf{27.24}  & \textbf{26.07}  & \textbf{24.88}  & \textbf{24.65}  & \textbf{24.19}  & \textbf{23.60}  & \textbf{23.24}  & \textbf{23.08}  & \textbf{22.61}  \bigstrut[b]\\
    \hline
    \hline
    \end{tabular}%
    }
\end{table*}

\subsection{Comparisons for Integer Scale Factors}

In this section, we take bicubic interpolation as the baseline and compare our proposed models with seven lightweight fixed-scale SR models: CARN \cite{ahn2018fast}, IMDN \cite{hui2019lightweight}, CTN \cite{wang2021contextual}, LatticeNet \cite{luo2020latticenet}, PAN \cite{zhao2020efficient}, CFSRCNN \cite{tian2020coarse}, and FeNet \cite{wang2022fenet}. All competitive models are retrained on each scale factor, while our proposed SalDRN is trained only once. In addition, for a fair comparison with existing fixed-scale SR models, we replace the LSUM in the proposed SalDRN with the same fixed-scale upsampling module in PAN to obtain a fixed-scale SR model named SalDRN-FS.

Table \ref{tab:results-fs} shows objective comparisons of SR results on three integer scale factors ($\times 2$, $\times 3$ and $\times 4$). We also compare the number of parameters of all models in Table \ref{tab:results-fs}. The SalDRN-FS achieves the best PSNR on all datasets and almost all scale factors, with up to $\SI{0.07}{\dB}$ improvement over the CFSRCNN on the GeoEye-1 dataset with a scale factor of $\times 2$. For the SSIM, SalDRN-FS achieves similar or better results than CFSRCNN, while the number of parameters is only $1/3$ of that of CFSRCNN. Compared with CTN with similar parameters, SalDRN-FS improves PSNR by up to $\SI{0.21}{\dB}$.

Our stepless SR model SalDRN performs slightly worse than SalDRN-FS because it is more challenging to learn SR for multiple scale factors simultaneously. Nevertheless, SalDRN still outperforms the competitive algorithms CARN, CTN, and PAN, obtaining objective metrics similar to IMDN. Note that SalDRN can achieve SR with arbitrary scale factors with only one model in the testing phase, so it can significantly save model training time and storage space. Therefore, SalDRN has more prominent advantages in model lightweighting.

\begin{figure}[t]
    \centering
    \includegraphics[width=\linewidth]{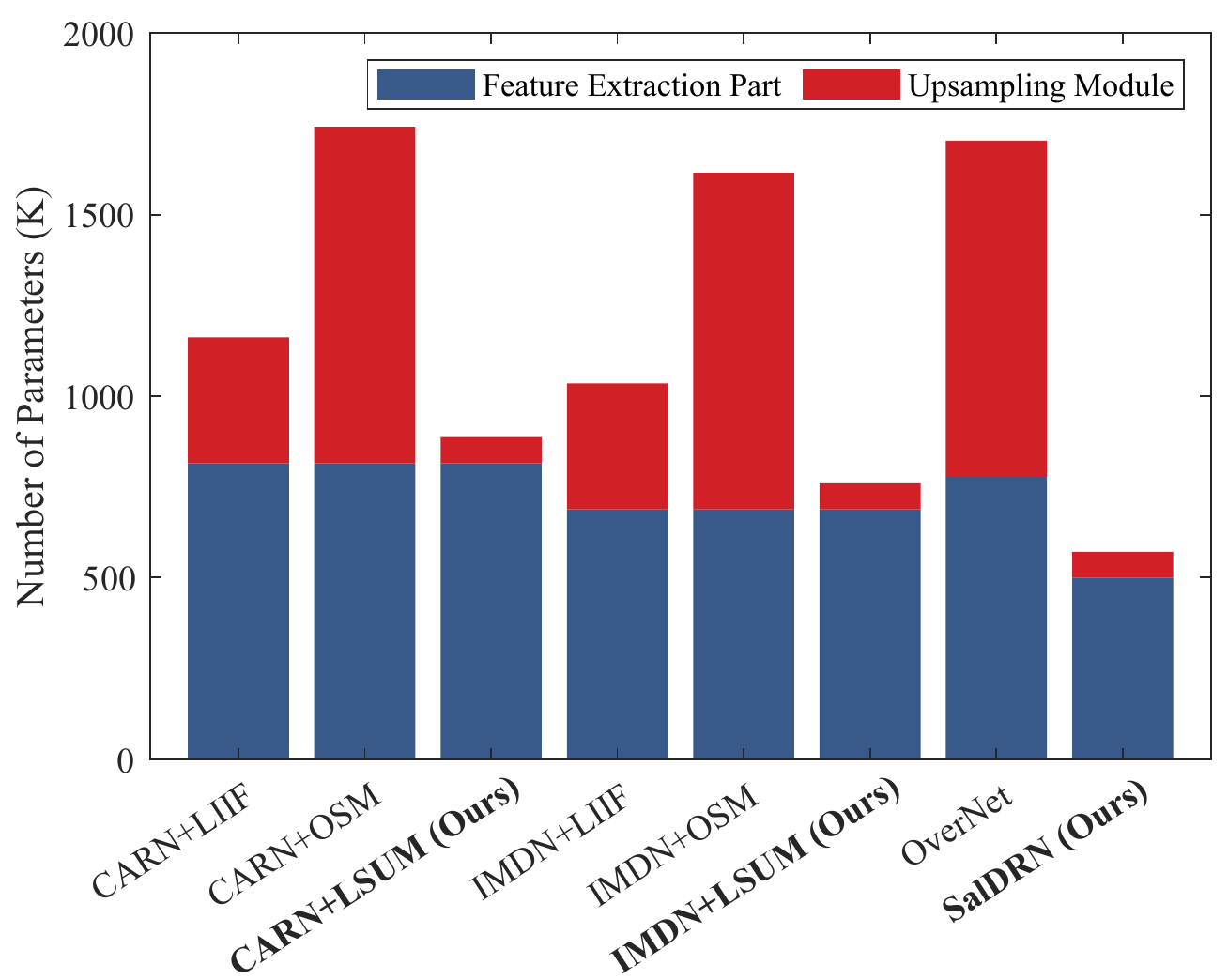}
    \caption{Parameter comparison of stepless SR models.}
    \label{fig:params2}
\end{figure}

\subsection{Comparisons for Arbitrary Scale Factors}

In this section, we compare the proposed SalDRN with competitive methods on the stepless SR task. Since the existing lightweight stepless SR is only OverNet \cite{behjati2021overnet}, to verify the effectiveness of the proposed LSUM, we construct comparison models by combining the lightweight feature extraction modules and the stepless upsampling modules. The feature extraction modules include CARN \cite{ahn2018fast} and IMDN \cite{hui2019lightweight}, and the stepless upsampling modules include LIIF \cite{chen2021learning}, OSM \cite{behjati2021overnet}, and our proposed LSUM. Therefore, we obtain six competitive methods: CARN+LIIF, CARN+OSM, CARN+LSUM, IMDN+LIIF, IMDN+OSM, and IMDN+LSUM. 

Fig.\,\ref{fig:params2} shows a parameter comparison of all competitive stepless SR models. The number of parameters of the LSUM is only $72\,\mathrm{K}$, which is $10\%$ of the feature extraction module in IMDN. Compared with the upsampling modules LIIF and OSM, the number of parameters is reduced by $80\%$ and $92\%$, respectively. Therefore, our proposed LSUM is an extremely lightweight upsampling module. 

Table \ref{tab:results-ms} lists the SR results of the SalDRN and seven competitive algorithms on the Google Earth and GeoEye-1 datasets. We randomly select twelve scale factors for display. Compared with the two upsampling modules proposed by previous works (LIIF and OSM), our proposed LSUM can achieve similar or better results with far fewer parameters. The SalDRN achieves the best performance on most scale factors due to the powerful feature expression capability of the DRM. 

\subsection{Model Complexity Analysis}

We evaluate the computational complexity of models from three aspects: number of  parameters, floating point operations (FLOPs), and runtime. The metric of FLOPs is defined as the number of multiplication and addition operations in the test phase averaged over image patches of $48\times 48$ on the Google Earth dataset. 

\begin{figure}[t]
    \centering
    \includegraphics[width=\linewidth]{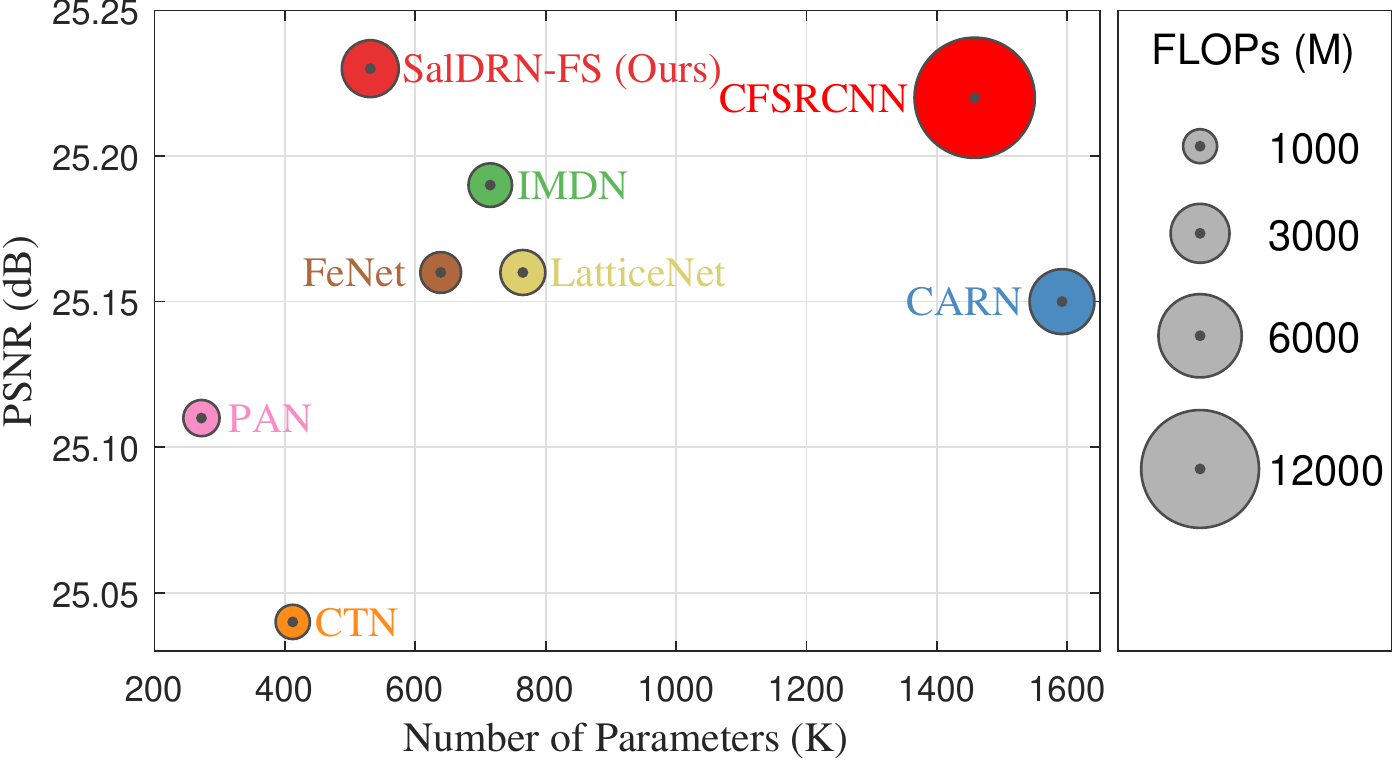}
    \caption{Comparison of PSNR, parameters, and FLOPs of fixed-scale models. Results are tested on the Google Earth dataset with a scale factor of $\times 4$.}
    \label{fig:parameter}
\end{figure}

Fig.\,\ref{fig:parameter} shows the performance, parameters, and FLOPs of SalDRN-FS and six lightweight models: CARN \cite{ahn2018fast}, IMDN \cite{hui2019lightweight}, CTN \cite{wang2021contextual}, LatticeNet \cite{luo2020latticenet}, PAN \cite{zhao2020efficient}, {CFSRCNN} \cite{tian2020coarse}, and FeNet \cite{wang2022fenet}. Our proposed SalDRN-FS achieves the best PSNR, and the number of parameters is only higher than the two super lightweight algorithms, PAN and CTN, ranking third. The FLOP of SalDRN-FS is $1225\,\mathrm{M}$, which is only $22\%$ of the closest performing model CFSRCNN. This performance is attributed to the dynamic routing strategy adopted by SalDRN-FS, which can reduce the computational complexity of the model while maintaining performance.

\begin{figure}[t]
    \centering
    \includegraphics[width=\linewidth]{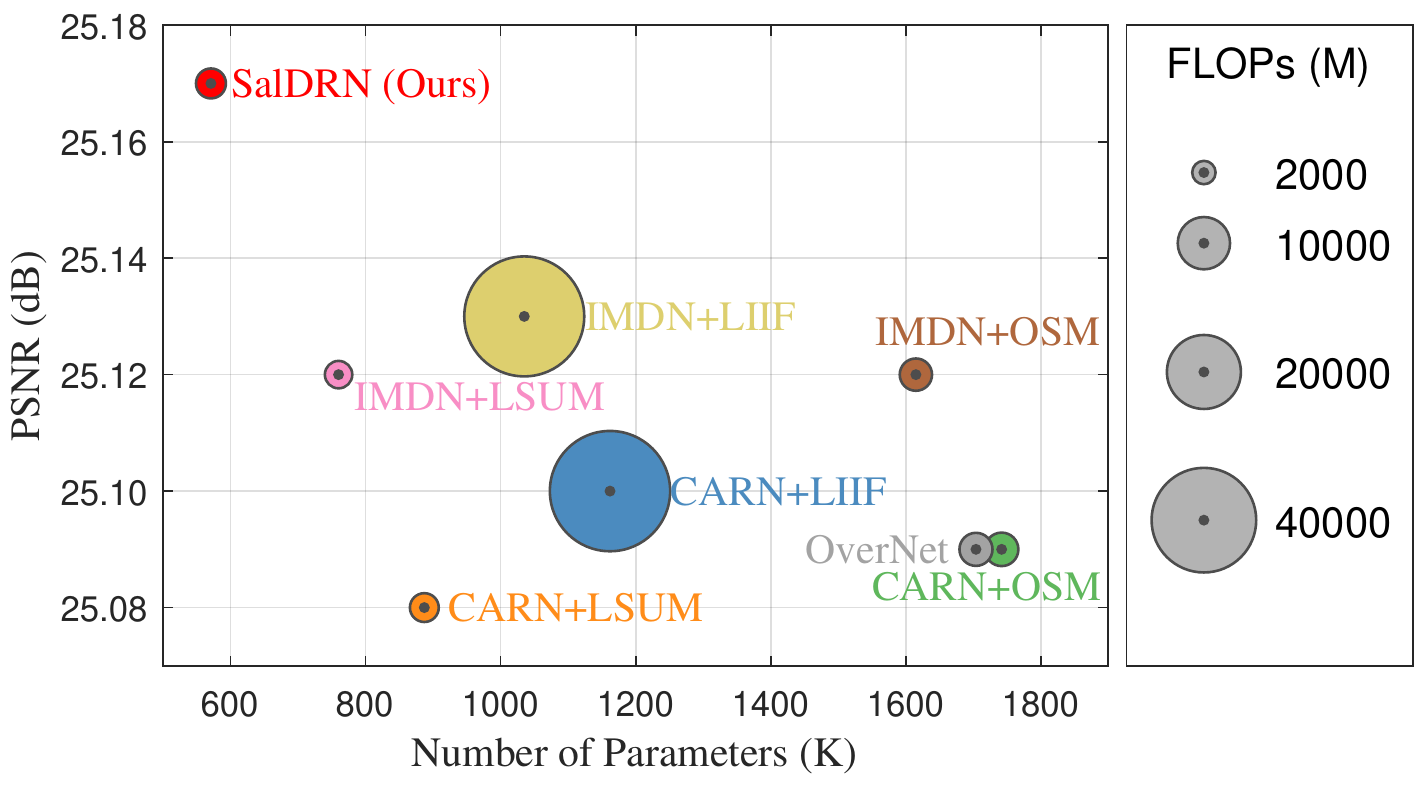}
    \caption{Comparison of PSNR, parameters, and FLOPs of stepless models. Results are tested on the Google Earth dataset with a scale factor of $\times 4$.}
    \label{fig:parameter-ms}
\end{figure}

\begin{figure*}[t]
    \centering
    \includegraphics[width=0.95\textwidth]{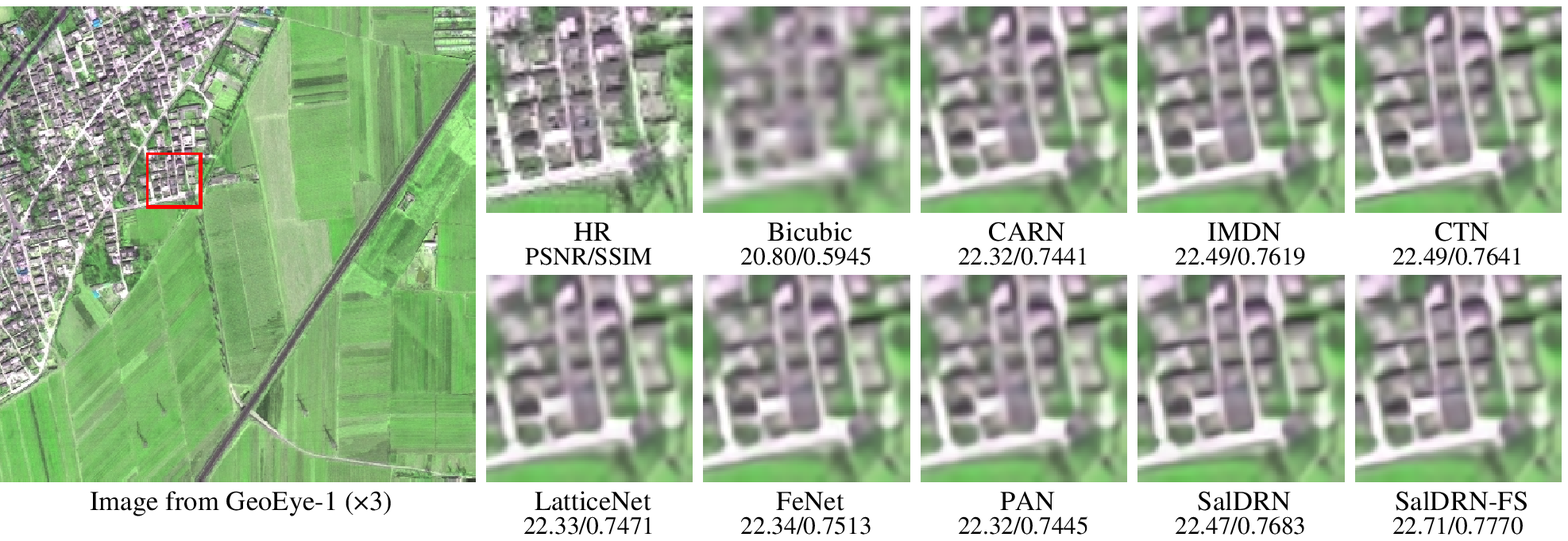}
    \vspace{4pt}\\
    \includegraphics[width=0.95\textwidth]{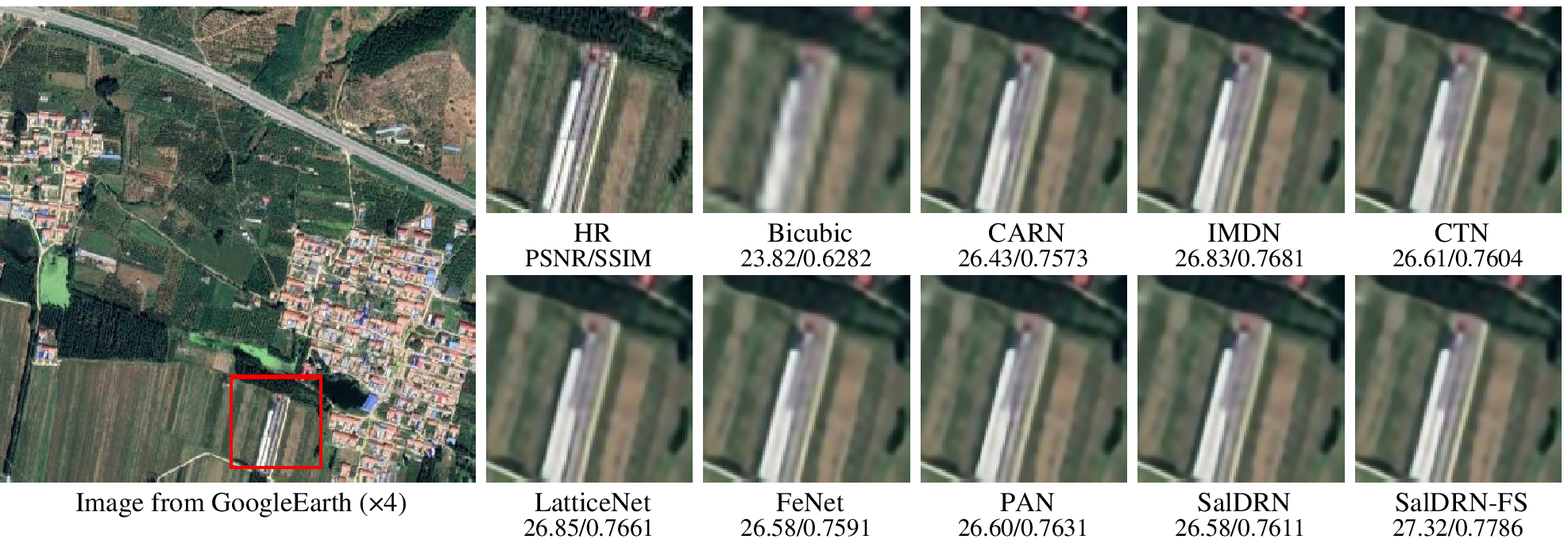}
    \caption{Visual comparisons with fixed-scale SR methods. The scale factors are $\times 3$ (top) and $\times 4$ (bottom), respectively.}
    \label{fig:mainrslt_fs}
\end{figure*}

\begin{figure*}[t]
    \centering
    \includegraphics[width=0.95\textwidth]{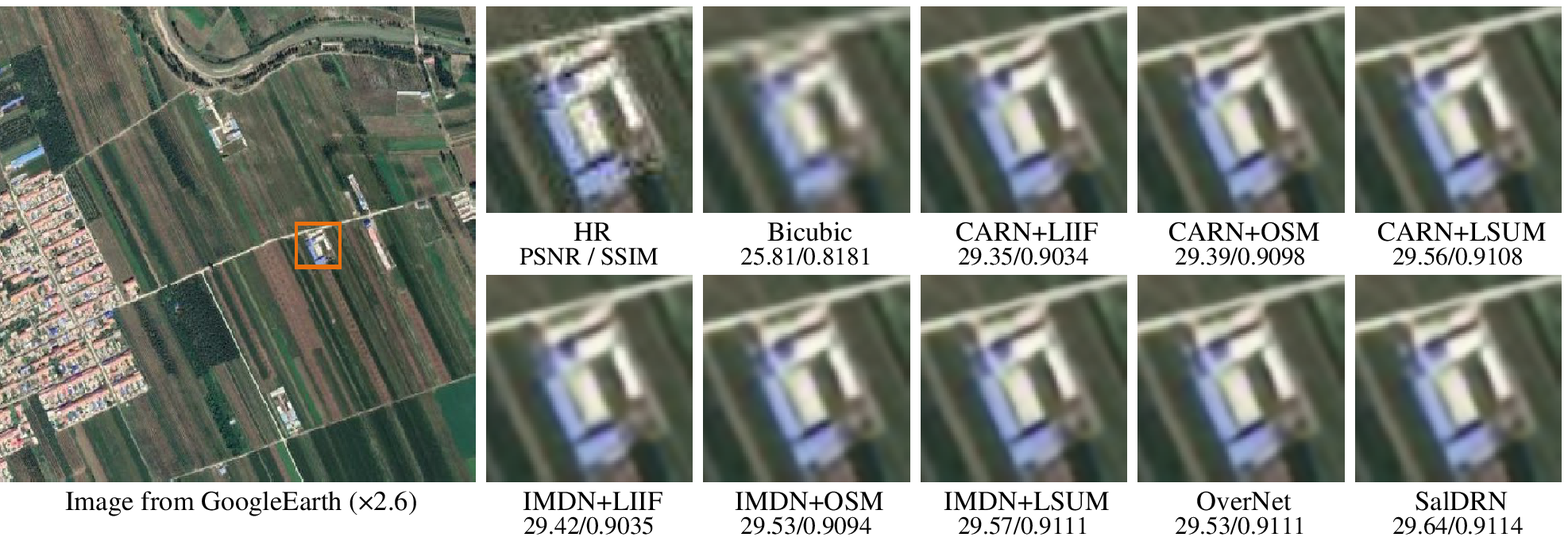}
    \vspace{4pt}\\
    \includegraphics[width=0.95\textwidth]{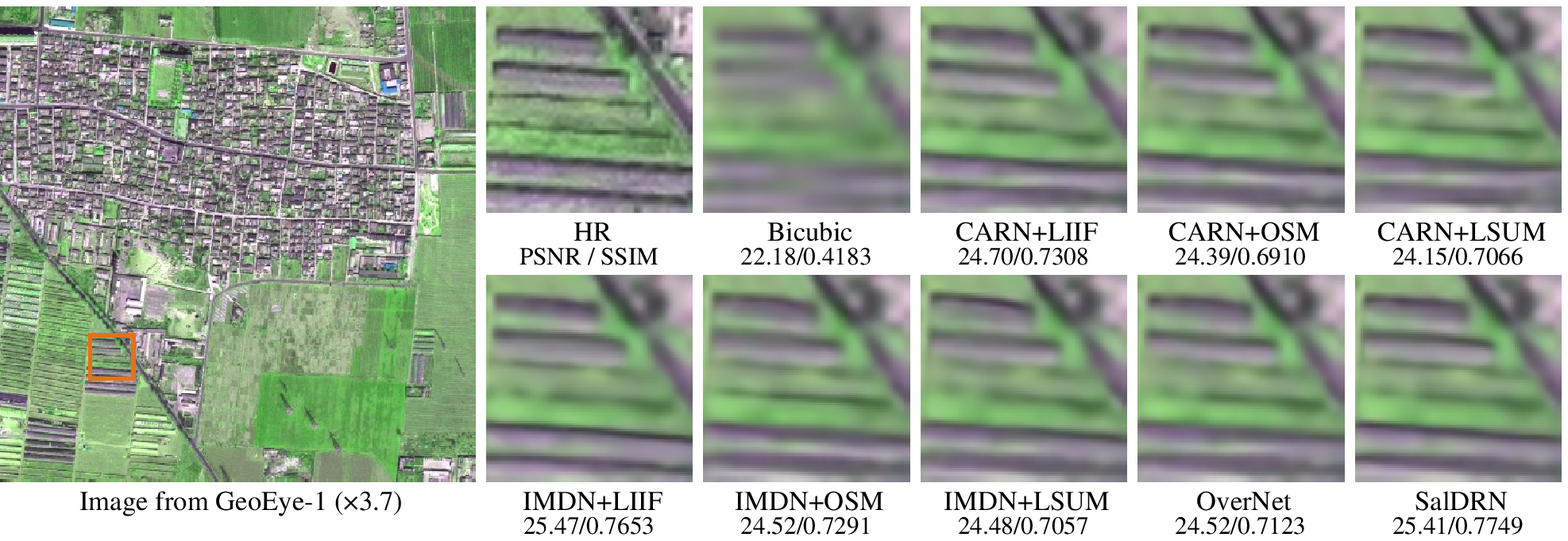}
    \caption{Visual comparisons with stepless SR methods. The scale factors are $\times 2.6$ (top) and $\times 3.9$ (bottom), respectively.}
    \label{fig:mainrslt_ms}
\end{figure*}

\begin{table}[t]
    \centering
    \renewcommand\arraystretch{1.2}
    \caption{Runtime comparison. The results are tested on the Google Earth dataset with a scale factor of $\times 4$. }
    \setlength{\tabcolsep}{2mm}{
        \begin{tabular}{cl|c}
            \hline
            \hline
            &  Methods & Runtime (ms) \bigstrut\\
            \hline
            \multicolumn{1}{c}{\multirow{7}[2]{*}{\makecell{Fixed-scale \\SR Models}}} & CARN \cite{ahn2018fast} & 4.6  \bigstrut[t]\\
            & IMDN \cite{hui2019lightweight} & 5.2  \\
            & LatticeNet \cite{luo2020latticenet} & 8.2  \\
            & PAN \cite{zhao2020efficient} & 8.3  \\
            & CFSRCNN \cite{tian2020coarse} & 11.4  \\
            & FeNet \cite{wang2022fenet} & 12.8 \\ 
            & SalDRN-FS (Ours) & 13.4  \\
            & CTN \cite{wang2021contextual} & 18.4  \bigstrut[b]\\
            \hline
            \multicolumn{1}{l}{\multirow{8}[0]{*}{\makecell{Stepless \\SR Models}}} & CARN\cite{ahn2018fast} + OSM\cite{behjati2021overnet} & 6.0  \bigstrut[t]\\
            & CARN\cite{ahn2018fast} + LSUM (Ours) & 6.5  \\
            & IMDN\cite{hui2019lightweight} + OSM\cite{behjati2021overnet} & 8.2  \\
            & IMDN\cite{hui2019lightweight} + LSUM (Ours) & 8.9  \\
            & OverNet\cite{behjati2021overnet} & 9.3  \\
            & SalDRN (Ours) & 15.4  \\
            & IMDN\cite{hui2019lightweight} + LIIF\cite{chen2021learning} & 218.3  \\
            & CARN\cite{ahn2018fast} + LIIF\cite{chen2021learning} & 221.6  \bigstrut[b]\\
            \hline
            \hline
        \end{tabular}%
    }
    \label{tab:runtime}
\end{table}

Fig.\,\ref{fig:parameter-ms} shows the performance, parameters, and FLOPs of SalDRN-FS and seven stepless SR models: CARN+LIIF, CARN+OSM, CARN+LSUM, IMDN+LIIF, IMDN+OSM, IMDN+LSUM, and OverNet. It can be seen that the models using LSUM not only have fewer parameters but also have more prominent advantages in the FLOPs. Compared with the upsampling modules LIIF and OSM, the FLOPs of LSUM are reduced by $98\%$ and $48\%$, respectively. Therefore, our proposed LSUM is one of the keys to constructing a lightweight stepless SR model. 

Table \ref{tab:runtime} shows a runtime comparison of all competitive algorithms. The results were tested on a workstation equipped with NVIDIA GeForce 3090 GPU, Intel 9700K CPU, and 64 GB memory. The test dataset was Google Earth, and the scale factor was $\times 4$. Among the fixed-scale SR models, the proposed SalDRN-FS runs faster than CTN and slightly slower than CFSRCNN, ranking sixth. Among all stepless SR methods, the models using LSUM run as fast as OSM and are more than $20$ times faster than the models using LIIF. Thanks to the LSUM and dynamic routing strategy, our proposal can better balance the contradiction between model performance and runtime. 

Comparing the performance, number of parameters, FLOPs, and runtime comprehensively, our proposed SalDRN and SalDRN-FS have lower model complexity and better SR performance. 

\subsection{Visual Comparisons}
We compare the visual results of our proposed methods with six fixed-scale SR models: CARN \cite{ahn2018fast}, IMDN \cite{hui2019lightweight}, CTN \cite{wang2021contextual}, LatticeNet \cite{luo2020latticenet}, PAN \cite{zhao2020efficient}, and FeNet \cite{wang2022fenet}. Fig.\,\ref{fig:mainrslt_fs} shows a visual comparison for scale factors of $\times 3$ and $\times 4$. The proposed SalDRN and SalDRN-FS can better recover edges, avoid road distortion, and generate more precise texture details.

Fig.\,\ref{fig:mainrslt_ms} shows a visual comparison of the SalDRN with seven stepless SR models on two non-integer scale factors. Models with the LSUM obtain comparable or even better visual results than previously proposed heavyweight upsampling modules. Our proposed SalDRN can obtain results in terms of texture clarity and edge sharpness.

\section{Discussion}

In this section, we first discuss the effectiveness of the dynamic routing strategy and the threshold settings in the DRM. Second, we study the effect of the parameter-sharing strategy in DRM. Third, we perform ablation experiments to verify the effectiveness of the SAPA mechanism in the LSUM. Finally, we verify the effectiveness of each term in the hybrid loss function.

\subsection{Study of Dynamic Routing Strategy}

\begin{figure*}[t]
    \centering
    \includegraphics[width=0.8\linewidth]{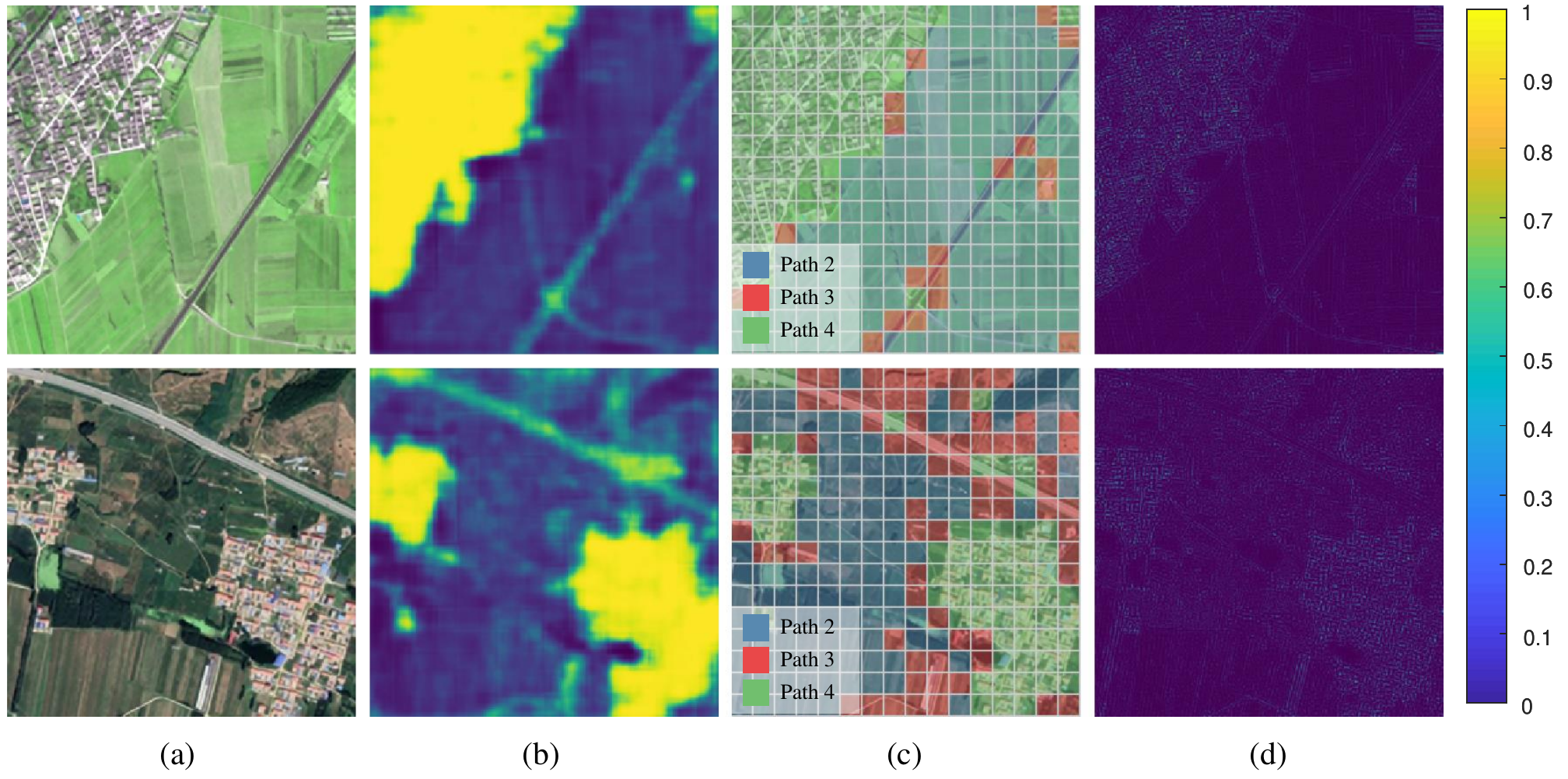}
    \caption{Visualization of the dynamic routing strategy. (a) SR results; (b) saliency maps; (c) visualization of path selection results; (d) error maps.}
    \label{fig:routing_vis}
\end{figure*}

To verify the effectiveness of the dynamic routing strategy, we visualize the output of the saliency detector and the path selection results in DRM. Fig.\,\ref{fig:routing_vis}\,(a) shows the SR results from the Google Earth and GeoEye-1 datasets. Fig.\,\ref{fig:routing_vis}\,(b) shows saliency maps output by the saliency detector. It can be seen that regions with complex textures have higher saliency values. Fig.\,\ref{fig:routing_vis}\,(c) shows the path selection results for each image patch. Since the threshold of the first path selection switch is set to $0$, only paths 2\,--\,4 are possible. Fig.\,\ref{fig:routing_vis}\,(d) shows the error maps of the SR results. The area with large errors indicates that the reconstruction is difficult. It can be seen that the saliency maps and error maps are consistent (positive correlation), indicating that the saliency map can be used as a guide for SR difficulty. Therefore, the dynamic routing strategy is reasonable and can indeed be used to improve the computational efficiency of SR models.

\subsection{Threshold Settings in DRM}

\begin{table}[ht]
    \centering
    \renewcommand\arraystretch{1.2}
    \caption{Comparison of different threshold settings with a scale factor of $\times 2$. The three values in the array represent the settings and results of the three path selection switches, respectively. ``\#'' represents the setting used in this article.}
    \label{tab:ablation-thresholds}
    \setlength{\tabcolsep}{1.5mm}{
        \begin{tabular}{l|lcc}
            \hline
            \hline
            Dataset & \multicolumn{3}{c}{GeoEye-1} \bigstrut\\
            \hline
            Threshold settings & Passing ratios & PSNR (\si{\dB}) & FLOPs (M) \bigstrut\\
            \hline
            [0, 0, 0] & [100\%, 100\%, 100\%] & 26.80  & 3410 (100\%) \bigstrut[t]\\
            {[0, 0.25, 0.5]} (\#) & {[100\%, 75\%, 47\%]} & 26.79  & 2554 (75\%) \\
            {[0, 0.25, 0.75]} & {[100\%, 75\%, 25\%]} & 26.78  & 2308 (68\%) \\
            {[0, 0.5, 0.75]} & [100\%, 47\%, 25\%] & 26.77  & 2008 (59\%) \\
            {[0, 0.75, 1]} & [100\%, 25\%, 0\%] & 26.72  & 1488 (44\%) \\
            {[0.25, 0.5, 0.75]} & [75\%, 47\%, 25\%] & 26.71  & 1730 (51\%) \\
            {[1, 1, 1]} & [0\%, 0\%, 0\%] & 24.96  & 117 (3\%) \bigstrut[b]\\
            \hline
            \hline
            Dataset & \multicolumn{3}{c}{Google Earth} \bigstrut\\
            \hline
            Threshold settings & Passing ratios & PSNR (\si{\dB}) & FLOPs (M) \bigstrut\\
            \hline
            {[0, 0, 0]} & [100\%, 100\%, 100\%] & 30.01  & 3410 (100\%) \bigstrut[t]\\
            {[0, 0.25, 0.5]} (\#) & [100\%, 94\%, 40\%] & 30.01  & 2677 (79\%) \\
            {[0, 0.25, 0.75]} & [100\%, 94\%, 18\%] & 30.01  & 2450 (72\%) \\
            {[0, 0.5, 0.75]} & [100\%, 39\%, 18\%] & 30.00  & 1833 (54\%) \\
            {[0, 0.75, 1]} & [100\%, 18\%, 0\%] & 29.96  & 1410 (41\%) \\
            {[0.25, 0.5, 0.75]} & [95\%, 19\%, 18\%] & 29.97  & 1775 (52\%) \\
            {[1, 1, 1]} & [0\%, 0\%, 0\%] & 28.20  & 117  (3\%) \bigstrut[b]\\
            \hline
            \hline
        \end{tabular}%
        }
\end{table}

We discuss the threshold settings of path select switches in DRM. Table \ref{tab:ablation-thresholds} shows the path selection ratio, PSNR, and FLOPs under different threshold settings. When the thresholds of the three path selection switches are set to $[0, 0, 0]$, only the deepest path in DRM is used for feature extraction, so it can be regarded as a model without the dynamic routing strategy. Under this setting, the best objective evaluation index can be obtained, but with the most FLOPs. When the thresholds are set to $[0, 0.25, 0.5]$, there is almost no reduction in PSNR, but the FLOPs on both datasets are reduced to $75\%$ and $79\%$, respectively. The dynamic routing strategy dynamically selects the feature extraction path according to the texture richness and reconstruction difficulty of the image patch, which can effectively reduce the computational complexity with little performance degradation. In addition, the FLOPs decrease as the threshold increases, but the model performance deteriorates to a certain extent. Therefore, we set the threshold as $[0, 0.25, 0.5]$ to trade off model performance and complexity.

\subsection{Discussion of Patch Size in DRM}

\begin{table}[t]
    \centering
    \renewcommand\arraystretch{1.2}
    \caption{Comparison of Different Patch Sizes in the DRM.  ``-/-'' indicates PSNR(\si{\dB}) and SSIM results.}
    \setlength{\tabcolsep}{1.4mm}{
      \begin{tabular}{c|c|c|c|c}
      \hline
      \hline
      \multirow{2}[2]{*}{Patch size} & \multicolumn{2}{c|}{GeoEye-1} & \multicolumn{2}{c}{Google Earth} \bigstrut \\
  \cline{2-5}          & $\times 2$    & $\times 4$    & $\times 2$    & $\times 4$ \bigstrut \\
      \hline
      32$\times$32 & 26.79/0.8613 & 22.41/0.6400 & 29.96/0.8809 & 25.16/0.6761 \bigstrut[t]\\
      40$\times$40 & 26.80/0.8616 & 22.39/0.6399 & 29.97/0.8811 & 25.15/0.6761 \\
      48$\times$48 & 26.80/0.8617 & 22.42/0.6408 & 29.97/0.8811 & 25.17/0.6768 \\
      56$\times$56 & 26.80/0.8618 & 22.37/0.6392 & 29.97/0.8812 & 25.14/0.6758 \\
      64$\times$64 & 26.80/0.8618 & 22.35/0.6383 & 29.97/0.8811 & 25.14/0.6759 \bigstrut[b]\\
      \hline
      \hline
      \end{tabular}%
    }
    \label{tab:ablation-patchsize}%
  \end{table}%

This section discusses the patch size that determines the SR difficulty. As shown in Table \ref{tab:ablation-patchsize}, we find that as the patch size increases, the receptive field increases, and the SR reconstruction results gradually improve. However, when the patch size becomes larger, the performance sometimes degrades at certain scale factors. We believe this occurs because the patch size is too large, and the saliency area in this patch has a small proportion, resulting in a low mean saliency value. This leads to this patch selecting a path with weak learning ability for feature extraction and being unable to fully mine the feature information. The accumulation of similar situations leads to a decrease in the evaluation metrics of the image SR reconstruction.

\subsection{Discussion of Parameter Sharing in DRM}
\label{sec:param_sharing}
\begin{table}[t]
    \centering
    \renewcommand\arraystretch{1.3}
    \caption{Discussion of parameter sharing in DRM.}
    \setlength{\tabcolsep}{2mm}{
        \begin{tabular}{c|cc|cc|cc}
            \hline
            \hline
            \multirow{2}[4]{*}{} & \multicolumn{2}{c|}{$\times 2$} & \multicolumn{2}{c|}{$\times 3$} & \multicolumn{2}{c}{$\times 4$} \bigstrut[t]\\
            \hline
            Parameter sharing  & $\checkmark$ & $\times$ & $\checkmark$ & $\times$ & $\checkmark$ & $\times$ \bigstrut\\\hline
            Params (K) & 519 & 910 & 519 & 910 & 530 & 921 \bigstrut[t]\\
            FLOPs (M) & 2554 & 1630 & 2772 & 1709 & 3031 & 1728 \\
            PSNR (dB) & 26.79  & 26.82  & 24.02  & 24.00  & 22.51  & 22.50  \bigstrut[b]\\
            \hline
            \hline
        \end{tabular}%
    }
    \label{tab:ablation-paramshare}
\end{table}

We discuss the parameter sharing strategy in the DRM. For this purpose, we construct a comparative model without parameter sharing named SalDRN-FS-w/o-PS. In SalDRN-FS-w/o-PS, the number of FRUs in the dynamic routing module is set to $3$. The number of IMDBs in the three FRUs is set to $3, 2, 2$ to avoid an excessively large number of parameters. Table \ref{tab:ablation-paramshare} shows a comparison of the SalDRN-FS and the SalDRN-FS-w/o-PS. We observe that the models achieve similar performance. SalDRN-FS-w/o-PS obtains a higher PSNR with a scale factor of $\times 2$, and the SalDRN-FS performs better with scale factors of $\times 3$ and $\times 4$. 

In terms of model complexity, SalDRN-FS has fewer parameters due to the parameter-sharing strategy but larger FLOPs. The result reflects a contradiction between the number of parameters, computational complexity, and model performance.

\subsection{Discussion of SAPA Mechanism}

\begin{figure*}[t]
    \centering
    \includegraphics[width=0.8\linewidth]{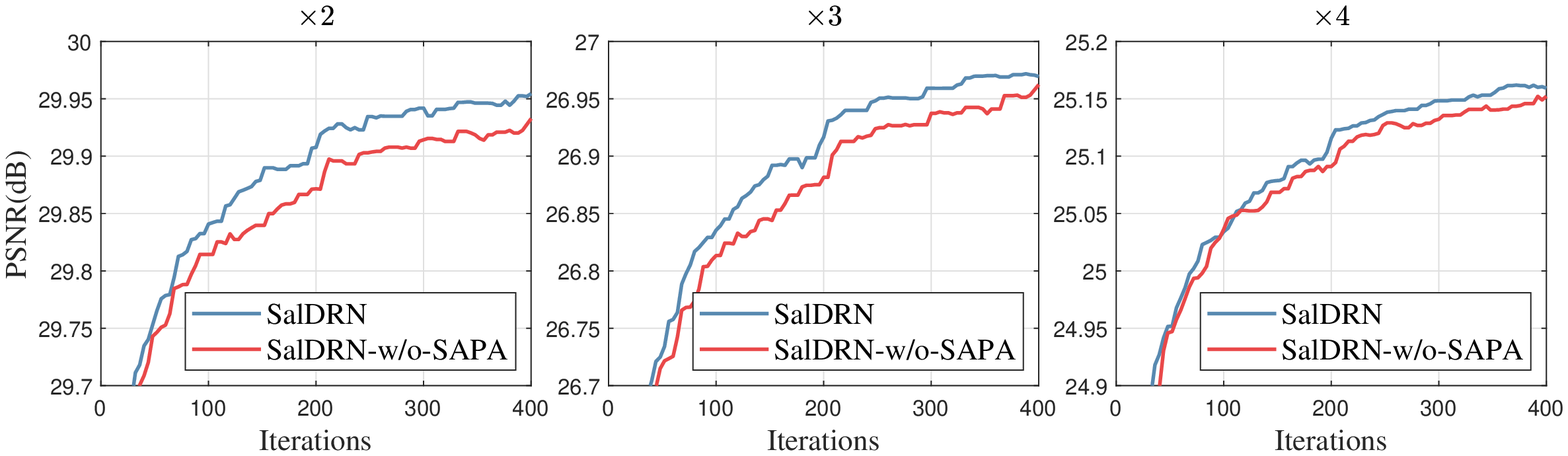}
    \caption{SAPA ablation experiment, PSNR tested on GeoEye-1 dataset.}
    \label{fig:ablation_sapa}
\end{figure*}

This section discusses the effectiveness of SAPA mechanisms. By removing the SAPA module in LSUM, a model without the scale-aware attention mechanism is obtained, named SalDRN-w/o-SAPA. Fig.\,\ref{fig:ablation_sapa} shows the convergence analysis of the SalDRN and SalDRN-w/o-SAPA on the GeoEye-1 dataset, where the curves represent the variation of PSNR with the number of iterations on three scale factors of $\times 2$, $\times 3$ and $\times 4$. As we can see, the SalDRN using SAPA has a faster convergence rate, and the PSNR is better than the comparison model without SAPA. The SAPA module can dynamically adjust the pixel-level attention weights according to the scale factor, thus enhancing the generalization performance over multiple scale factors.

\begin{figure*}[t]
    \centering
    \includegraphics[width=0.8\linewidth]{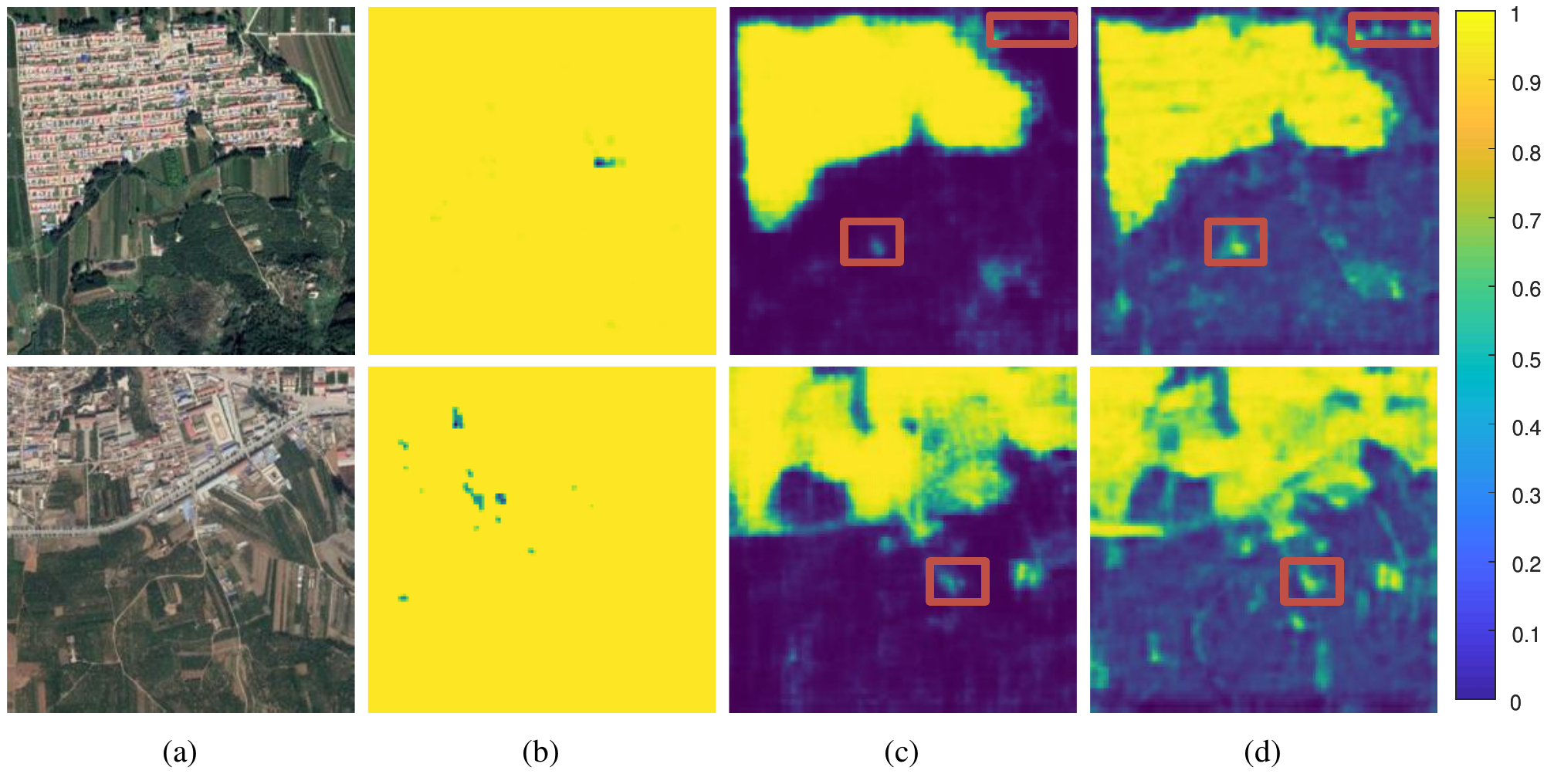}
    \caption{Comparison of saliency maps generated by models using different loss functions. (a) Input images; (b) Model-I; (c) Model-II; (d) SalDRN. }
    \label{fig:ablation-loss}
\end{figure*}

\subsection{Discussion of Hybrid Loss Function}

\begin{table}[t]
    \centering
    \renewcommand\arraystretch{1.2}
    \caption{Comparison of PSNR(\si{\dB}) with different loss function settings.}
    \label{tab:ablation-loss}
    \setlength{\tabcolsep}{1mm}{
    \begin{tabular}{c|ccc|ccc|ccc}
        \hline
        \hline
        \multirow{2}[3]{*}{Models} & \multirow{2}[3]{*}{$\mathcal L_{\mathrm{SR}}$} & \multirow{2}[3]{*}{$\mathcal L_{\mathrm{sal}}$} & \multirow{2}[3]{*}{\makecell{$\mathcal L_{\mathrm{diff}}$}} & \multicolumn{3}{c|}{GeoEye-1} & \multicolumn{3}{c}{Google Earth} \bigstrut\\
        \cline{5-10}   &    &    &    & $\times 2$ & $\times 3$ & $\times 4$ & $\times 2$ & $\times 3$ & $\times 4$ \bigstrut\\
        \hline
        Model-I & $\checkmark$ & $\times$ & $\times$ & 26.80  & 23.96  & 22.41  & 29.99  & 27.00  & 25.18  \bigstrut[t]\\
        Model-II & $\checkmark$ & $\checkmark$ & $\times$ & 26.78  & 23.96  & 22.41  & 29.95  & 26.97  & 25.16  \\
        SalDRN & $\checkmark$ & $\checkmark$ & $\checkmark$ & 26.80  & 23.98  & 22.42  & 29.97  & 26.97  & 25.17  \bigstrut[b]\\
        \hline
        \hline
    \end{tabular}%
    }
\end{table}

In this section, we discuss the effectiveness of each term in the hybrid loss function, SR loss $\mathcal{L}_{\mathrm{SR}}$, saliency loss $\mathcal{L}_{\mathrm{sal}}$, and SR difficulty loss $\mathcal{L}_{\mathrm{diff}}$. We add these three items one by one to obtain three comparison models. The model trained with only the SR loss is denoted Model-I; the model trained with the SR loss and the saliency loss is denoted Model-II. Table \ref{tab:ablation-loss} shows a performance comparison between SalDRN, Model-I, and Model-II on the GeoEye-1 and Google Earth datasets. 

As seen from Table \ref{tab:ablation-loss}, the model with only SR loss achieves the highest objective metrics. This is because if no constraints are imposed on the saliency detector, it will output a saliency map with all pixels as $1$ (as shown in Fig.\,\ref{fig:ablation-loss}\,(b)), so that the most image patches can pass the deepest path. Therefore, only using the SR loss will completely invalidate the dynamic routing strategy. If the SR difficulty loss is removed, then the performance of Model-II decreases slightly. This is because the saliency map only concentrates on regions of visual interest, such as residential areas, while ignoring other areas with greater SR difficulty, as shown in Fig.\,\ref{fig:ablation-loss}\,(c) and (d). Since SalDRN adopts the SR difficulty loss, it can better guide the network for path selection and obtain better objective metrics.
\section{Conclusions}
\label{sec:conclusion}

In this study, to improve the performance of SR in scenarios with limited computing resources and meet the needs of arbitrary scale factor amplification, we proposed a lightweight SalDRN for stepless SR of RSIs. We designed a super-lightweight saliency detector to embed in the SR framework with a negligible additional computational burden. The proposed dynamic routing strategy crops the input image into small patches and selects the appropriate deep network branch for feature extraction according to the SR difficulty. We also proposed an LSUM for lightweight SR networks based on implicit feature functions and the pixel attention mechanism. LSUM enables the network to process multiple scale factors within a single model, significantly reducing the training and storage costs. Experimental results showed that the proposed SalDRN narrows the performance gap with heavyweight SR networks compared with SOTA lightweight SR models.

\balance


%

\appendices



\ifCLASSOPTIONcaptionsoff
    \newpage
\fi



\bibliographystyle{IEEEtran}
\bibliography{IEEEabrv,refs}

\end{document}